\documentclass{article}

\usepackage{microtype}
\usepackage{graphicx}
\usepackage{subcaption}
\usepackage{booktabs} 

\usepackage{hyperref}
\usepackage{float} 
\usepackage{xcolor}
\newcommand{\red}[1]{\textcolor{red}{#1}}
\newcommand{\blue}[1]{\textcolor{blue}{#1}}
\newcommand{\gray}[1]{\textcolor{black!70}{#1}}

\usepackage[accepted]{icml2026}

\usepackage{amsmath}
\usepackage{amssymb}
\usepackage{mathtools}
\usepackage{amsthm}

\usepackage{makecell}
\usepackage{multirow}
\usepackage[capitalize,noabbrev]{cleveref}
\usepackage[table]{xcolor}
\theoremstyle{plain}

\theoremstyle{definition}

\theoremstyle{remark}

\usepackage[textsize=tiny]{todonotes}
\newcommand{\myparagraph}[1]{\vspace{1.5pt}\noindent\textbf{#1}}
\usepackage{xspace}
\usepackage{wrapfig}
\usepackage{fontawesome5}
\icmltitlerunning{Mitigating Mask Prior Drift and Positional Attention Collapse in LDVLMs}
\begin{document}

\twocolumn[
  \icmltitle{Mitigating Mask Prior Drift and Positional Attention Collapse \\ in Large Diffusion Vision-Language Models}

  \begin{icmlauthorlist}
    \icmlauthor{Sujung Hong}{yonsei}
    \icmlauthor{Chanyong Yoon}{yonsei}
    \icmlauthor{Seong Jae Hwang}{yonsei}
  \end{icmlauthorlist}

  \icmlaffiliation{yonsei}{Department of Artificial Intelligence, Yonsei University, Seoul, Republic of Korea}

  \icmlcorrespondingauthor{Seong Jae Hwang}{seongjae@yonsei.ac.kr}
  
    \begin{center}
        \faGithub\ \href{https://github.com/hong-sujung/MPD-PAC}{\texttt{hong-sujung/MPD-PAC}}
        \quad
        \faGlobe\ \href{https://hong-sujung.github.io/MPD-PAC}{Project Page}
    \end{center}
  \icmlkeywords{Machine Learning, ICML}

  \vskip 0.3in
]



\printAffiliationsAndNotice{}  

\begin{abstract}
Large diffusion vision–language models (LDVLMs) have recently emerged as a promising alternative to autoregressive models, enabling parallel decoding for efficient inference and leveraging bidirectional attention for global context.
Despite these advances, their behavior under long-form generation remains underexplored.
In this work, we show that existing LDVLMs suffer from repetitive generation and degraded visual grounding, and identify two underlying causes.
First, repetitive generation originates from a mask token prior: since generation tokens are initialized as mask tokens, their hidden representations progressively drift toward a shared prior direction over generation steps.
Second, a fundamental misalignment between the positional attention bias and the iterative unmasking process suppresses attention toward informative visual tokens, degrading visual grounding.
Based on these insights, we propose a training-free approach, introducing Mask Prior Suppression and Monotonic RoPE Scaling to mitigate mask prior drift and positional attention collapse during decoding.
Experiments on general multimodal benchmarks and visual grounding tasks demonstrate improvements over baseline LDVLMs, with robust gains on long-form description benchmarks.
Our results show that these failures can be effectively addressed with a lightweight, plug-and-play strategy that requires no additional training and generalizes across diverse LDVLM architectures.
\end{abstract}

\section{Introduction}
Autoregressive large language models (LLMs) have become the dominant paradigm for reasoning and natural language generation~\cite{brown2020language,radford2019language,dubey2024llama,touvron2023llama,peng2023instruction,wei2022chain}.
However, their sequential decoding limits inference efficiency and prevents refinement~\cite{khoshnoodi2024comprehensive}.
To address these limitations, large language diffusion models (LLDMs)~\cite{ye2025dream,nie2025large,li2025survey} have emerged as a promising alternative.
Building upon this paradigm, diffusion-based generation has been extended to the vision-language domain, resulting in large diffusion vision-language models (LDVLMs)~\cite{li2025LaViDa,yang2025mmada,you2025llada}.
These models enable globally consistent multimodal reasoning by maintaining interactions between visual and textual tokens throughout the generation process.

LDVLMs exhibit several distinctive advantages over autoregressive multimodal models.
First, LDVLMs enable explicit control over the trade-off between generation quality and inference latency by adjusting the number of generation steps.
They employ an iterative unmasking process, in which all generation tokens are initialized from a shared mask token $\mathcal{M}$ and progressively refined into meaningful tokens through a fixed number of generation steps.
Second, LDVLMs facilitate global context modeling and structured output formation, which are critical for complex vision--language tasks.
They adopt a bidirectional unmasking mechanism that allows each token to attend to all other tokens at every generation step, in contrast to the causal attention imposed by autoregressive models.
To support stable bidirectional attention, LDVLMs typically employ Rotary Position Embeddings (RoPE)~\cite{roformer}, which encode relative positional information while preserving access to global context.

\begin{figure}[t]
    \centering
    \includegraphics[width=\linewidth]{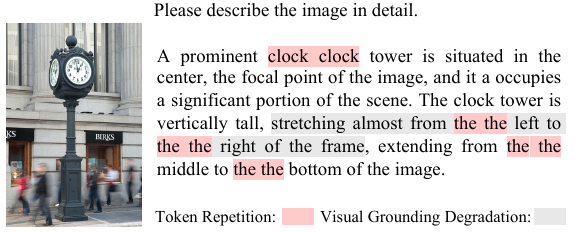}
    \caption{
    \textbf{Failure case of LDVLMs.}
        Under parallel decoding with 64 generation tokens and 16 generation steps,
        LLaDA-V produces highly repetitive phrases as highlighted in \red{red},
        and exhibits degraded visual grounding as highlighted in \gray{gray}.
    }
    \label{fig:fig_motivation}
    \vspace{-10pt}
\end{figure}

Despite these strengths, the behavior of LDVLMs in complex multimodal settings remains insufficiently explored.
In this work, we identify two fundamental challenges that arise during long-form multimodal generation,
as illustrated in~\cref{fig:fig_motivation}.
First, we observe a persistent repetition of specific tokens during decoding that is largely independent of the input text.
This phenomenon becomes increasingly severe as the number of generation steps decreases.
We refer to this behavior as \textit{mask prior drift}.
Because all generation tokens are initialized from the same mask token $\mathcal{M}$,
their hidden representations progressively converge toward a shared prior direction.
This convergence limits semantic diversity across tokens and ultimately leads to repetitive generation.
Second, we find that LDVLMs suffer from degraded visual grounding in long-form generation.
Our analysis attributes this issue to positional attention collapse, induced by the inherent locality bias of RoPE.
During iterative unmasking, attention concentrates on nearby semantically incomplete mask tokens, suppressing attention to distant informative visual tokens.
Together, these issues weaken visual--textual alignment
and limit the reliability of LDVLMs in producing high-quality long-form descriptions.

To address these issues, we propose two inference-time techniques: Mask Prior Suppression and Monotonic RoPE Scaling.
First, Mask Prior Suppression reduces representational drift by adjusting the final hidden states of generation tokens in the mask prior subspace.
This operation effectively reduces structured repetition while preserving semantic expressiveness.
Second, we introduce Monotonic RoPE Scaling to improve positional attention stability.
By monotonically emphasizing low-frequency RoPE components, generation tokens maintain stable attention to distant visual tokens and preserve RoPE's relative distance structure.
Both techniques operate entirely at inference time and require no parameter updates or retraining.

Our main contributions are summarized as follows.
First, we present a systematic analysis of the stability of the generation in LDVLMs, revealing repetitive patterns and degraded visual grounding as significant challenges. 
Second, we propose two training-free techniques, Mask Prior Suppression and Monotonic RoPE Scaling, that directly address the representational and positional biases underlying these failures.
Finally, extensive experiments across diverse benchmarks and LDVLM architectures demonstrate consistent improvements in visual grounding and long-form generation quality, without retraining or additional parameters.
Together, our methods provide a principled and lightweight solution for stabilizing long-form multimodal generation in LDVLMs.
\section{Preliminaries}
\subsection{Large Diffusion Language Model}
\vspace{-5pt}
Diffusion Large Language Models (DLLMs) generate text via an iterative denoising procedure with parallel decoding, offering an alternative to autoregressive generation.
Among these methods~\cite{austin2021structured, sahoo2024simple, lou2023discrete, gong2024scaling, nie2024scaling, nie2025large, ye2025dream}, Masked Diffusion Models (MDMs) have emerged as a representative framework for discrete sequence modeling.
MDMs assume an input sequence $x_0 = [x^i]^N_{i=1}$ consisting of $N$ tokens, including special mask tokens $\mathcal{M}$.
The model defines the model distribution $p_\theta(x_0)$ through a forward and reverse diffusion process.
In the forward process, $x_0$ is independently replaced by $\mathcal{M}$ with the time step $t$ uniformly sampled from the interval $[0,1]$.
The reverse process starts from a fully masked sequence and iteratively replaces
$\mathcal{M}$ tokens to reconstruct the original sentence by sampling token values
at masked positions from the conditional distribution
$p_\theta(x_0 \mid x_t)$.
During inference, generation proceeds for a fixed number of reverse steps $T$,
where all masked positions are updated in parallel at each step.
The number of steps $T$ governs the trade-off between decoding efficiency and
generation quality.

\subsection{Rotary Position Embedding}
\vspace{-5pt}

\begin{figure*}[t]
    \centering
    \includegraphics[width=\linewidth]{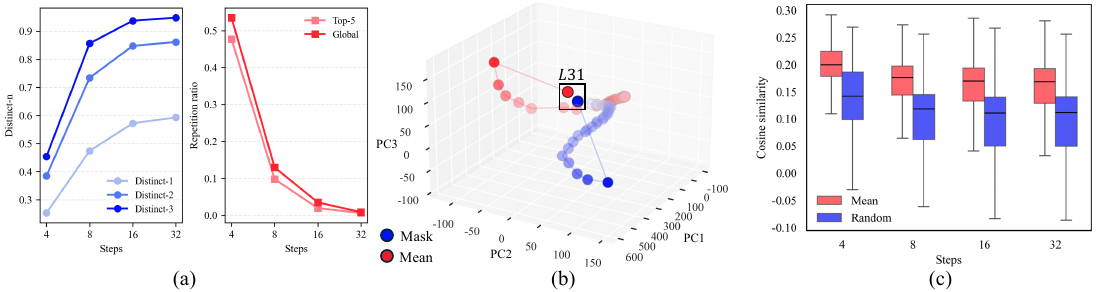}
    \caption{
    \textbf{Visualization of token repetition and mask prior drift.}
    (a) Distinct-$n$ (left) and repetition ratio (right) across different numbers of generation steps.
    Fewer generation steps lead to lower distinct-$n$ and higher repetition.
    (b) 3D PCA trajectories of hidden states for the vocabulary mean embedding and the uncontextualized mask token, which converge to a similar region at the final layer ($L31$).
    (c) Cosine similarity between contextualized mask token embeddings and the vocabulary mean,
    showing consistently stronger alignment than random embeddings, especially with fewer generation steps.
    }
    \label{fig:analysis1}
    \vspace{-10pt}
\end{figure*}
Rotary Position Embedding (RoPE)~\cite{roformer} is a widely used mechanism for encoding
relative positional information in transformer-based models.
Let $x_m \in \mathbb{R}^d$ denote the $d$-dimensional embedding vector at position $m$,
where $d$ is even.
The query and key representations are obtained via linear projections
$q_m = W_Q x_m$ and $k_n = W_K x_n$,
where $W_Q, W_K \in \mathbb{R}^{d \times d}$.
RoPE encodes positional information by partitioning the embedding dimensions into
$d/2$ two-dimensional subspaces.
Each subspace is associated with a frequency
$\theta_i = 10000^{-2i/d}$ for $i \in \{0, 1, \ldots, d/2 - 1\}$,
and applies a position-dependent rotation
\[
r(\theta_i, m) =
\begin{pmatrix}
\cos(m\theta_i) & -\sin(m\theta_i) \\
\sin(m\theta_i) & \cos(m\theta_i)
\end{pmatrix}.
\]
Stacking all subspaces yields a block-diagonal rotation matrix
$R_{\Theta,m}^d = \mathrm{diag}\bigl(r(\theta_0,m), \ldots, r(\theta_{d/2-1},m)\bigr)$,
which is applied to the query and key vectors as
$q'_m = R_{\Theta,m}^d q_m$ and $k'_n = R_{\Theta,n}^d k_n$.
A key property of RoPE is that the resulting attention score depends only on the
relative position between tokens, i.e.,
${q'}_m^\top k'_n = q_m^\top R_{\Theta,n-m}^d k_n$.
Thus, RoPE implicitly encodes relative positional information through rotational
offsets in the query--key interaction.

\section{Related Work}
\subsection{Large Diffusion Vision--Language Model}
\vspace{-5pt}
DLLMs have recently been extended to the vision--language domain.
Previous studies mainly focus on designing training frameworks with multimodal data~\cite{you2025llada,li2025LaViDa,yang2025mmada}.
For instance, LaViDa~\cite{li2025LaViDa} introduces complementary masking, which ensures that all tokens are considered across masking patterns during training.
Similarly, MMaDA~\cite{yang2025mmada} and Lumina-DiMOO~\cite{xin2025lumina} proposes a unified architecture supporting textual reasoning, multimodal reasoning, and text-to-image generation.
A complementary line of work targets the decoding process itself, for example by reducing redundant suffix-mask attention~\cite{dpad} or by restructuring the decoding paradigm into block-wise variants~\cite{d2f}.
These directions modify the structural inputs or schedule of unmasking, whereas our method operates at the representation and positional levels and is largely orthogonal to such structural interventions; we provide a direct empirical comparison in~\cref{sec:app_dpad}.
Despite these advances, prior work has paid limited attention to intrinsic generation characteristics in LDVLMs for visual--textual alignment.
\vspace{-5pt}

\subsection{Rotary Position Embeddings in VLMs}
\vspace{-5pt}
Rotary Position Embedding (RoPE) has been extended to multimodal models through two broad approaches.
The first approach applies the one-dimensional positional mechanism originally designed for text tokens to visual tokens, often combined with dynamic position scaling to handle large positional indices~\cite{roformer,ge2025v2pe}.
The second approach assigns distinct frequency components to different spatial or temporal dimensions to better capture multimodal structure~\cite{wang2024qwen2,wei2025videorope,li2025hope}.
In addition, CircleRoPE~\cite{wang2025circle} arranges visual tokens in a circular layout orthogonal to text, promoting uniform cross-modal attention.
While prior work extends positional encoding for multimodal autoregressive models, typically requiring training, positional mechanisms for MDMs with bidirectional attention and parallel unmasking remain largely unexplored.
\vspace{-5pt}
\begin{figure*}[!t]
\includegraphics[width=\textwidth]
{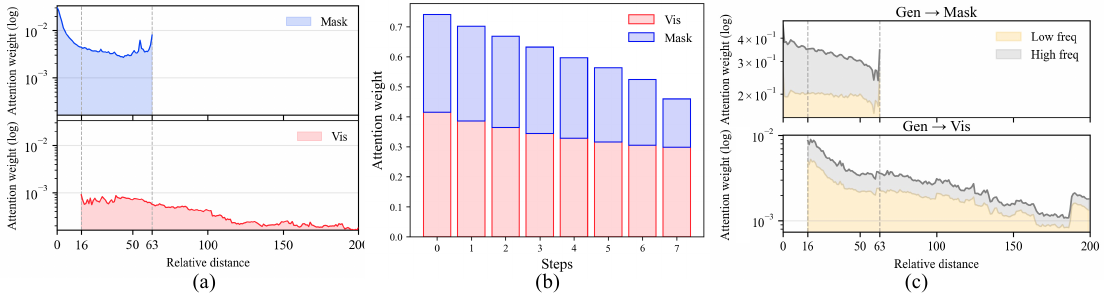}
\caption{
\textbf{Visualization of positional attention collapse.}
(a) Mean attention weight (log scale) across relative distance, showing stronger attention to mask tokens than visual tokens at similar distances and an overall decreasing trend in attention to visual tokens as relative distance increases.
(b) Sum of attention to visual and mask tokens per generation token across generation steps, revealing a persistent allocation of comparable attention weights to mask tokens despite their lack of semantic content.
(c) Frequency decomposition of attention over relative distance, where RoPE dimensions are evenly divided into high- and low-frequency bands.
High-frequency components dominate at short ranges, while low-frequency components dominate at long ranges, yet overall attention remains weak.
}
\label{fig:freq}
\vspace{-10pt}
\end{figure*}
\section{Analysis of Generation Failures in LDVLMs}
\label{sec:analysis}
While LDVLMs demonstrate competitive performance relative to autoregressive models, they often suffer from repetitive generation patterns and suboptimal visual grounding.
In this section, we investigate the causes of these behaviors by analyzing the underlying architectural design and hidden state dynamics.
\vspace{-5pt}

\subsection{Token Repetition and Mask Prior Bias}
\vspace{-5pt}
\textbf{Speed--quality trade-off and repetition.}
LDVLMs exhibit an inherent trade-off between decoding speed and generation quality, which is closely related to the number of tokens unmasked at each iteration.
As the number of generation steps decreases, a larger proportion of tokens is updated simultaneously.
This parallel unmasking reduces the effective conditioning depth through which contextual information can be propagated across transformer layers.
In practice, such behavior frequently manifests as consecutive token repetition during inference.
Although this speed--quality trade-off has been widely observed, the underlying causes of quality degradation, particularly repetitive generation,  remain insufficiently understood.

\vspace{-3pt}
\textbf{Structured token repetition.}
To analyze this phenomenon, we perform multimodal text generation on 100 randomly sampled images from the COCO 2014 validation set using the instruction ``\texttt{Please describe the image in detail}.''
For each image, we generate 64 tokens while varying the number of generation steps $T\in\{4, 8, 16, 32\}$.
We evaluate generation quality using two metrics: \textit{distinct-$n$} and \textit{repetition ratio}.
Distinct-$n$ measures the proportion of unique $n$-grams to assess lexical diversity, while repetition ratio quantifies the fraction of consecutively repeated tokens.
As shown in~\cref{fig:analysis1}(a, left), decreasing the number of steps leads to a sharp decline in distinct-n.
Importantly, we find that this repetition is highly structured rather than random, concentrating on a small set of function tokens.
By analyzing the top-$k$ logit tokens of the uncontextualized mask token
(e.g., \texttt{\textbackslash n}, \texttt{ the}, \texttt{,}, \texttt{<space>}, \texttt{.}), we observe that these few tokens account for over 95\% of all repeated outputs, as illustrated in~\cref{fig:analysis1}(a, right).
Here, an \textit{uncontextualized mask token} refers to the mask token processed without any surrounding text or visual input.
This result suggests that as the number of generation steps decreases, the model fails to leverage contextual signals and instead collapses toward the prior encoded in the mask token.

\vspace{-3pt}
\textbf{Mask prior drift in hidden states.}
We hypothesize that this structured repetition originates from the shared initialization of all generation tokens as the mask token, which introduces a representation prior into the decoding process.
The mask token is trained to represent a context-agnostic and uninformative state, and its embedding is therefore biased toward the mean of the vocabulary embedding space.
Let $f_l(\cdot)$ denote the transformation induced by the $l$-th transformer layer, and let $h_l^{\mathcal{M}} = f_l(e_{\mathcal{M}})$ be the hidden state of the mask token at layer $l$.
We define the vocabulary embedding mean as $\hat{e} = \frac{1}{|V|}\sum_{v \in V} e_v$, where $V$ denotes the set of vocabulary embeddings, and similarly obtain its layer-wise representations as $h_l^{\hat{e}} = f_l(\hat{e})$.
As shown in~\cref{fig:analysis1}(b), a 3D PCA visualization of the joint set
$\{ h_l^{\mathcal{M}} \}_{l=1}^{L} \;\cup\; \{ h_l^{\hat{e}} \}_{l=1}^{L}$, where $L$ denotes the total number of transformer layers, reveals that the hidden states $h_l^{\mathcal{M}}$ and  $h_l^{\hat{e}}$  progressively align at the final transformer layer.
While their trajectories remain closely aligned in the early layers and exhibit distinct characteristics in the intermediate layers, they converge toward a similar region in the final layer ($L31$).

\vspace{-3pt}
\textbf{Mask prior drift during inference.}
We further investigate whether this representation drift persists when mask tokens become contextualized during inference.
Here, contextualized refers to tokens that are processed jointly with other visual and textual tokens as part of the model input.
We use the same dataset and instructions as in the previous experiment and perform generation on 10 randomly sampled images, producing 64 tokens for each image.
We find that contextualized mask token embeddings exhibit higher cosine similarity to the vocabulary mean embedding than randomly sampled token embeddings, as illustrated in~\cref{fig:analysis1}(c).
This effect is exacerbated as the number of decoding steps decreases, since parallel decoding restricts conditioning signals and amplifies the bias of the mask token, ultimately leading to highly structured token repetition.

\textbf{Empirical scope of our claim.}
We emphasize that this analysis is empirical: we do \emph{not} claim that the vocabulary mean fully characterizes the nonlinear dynamics of mask-token collapse.
Rather, motivated by the Linear Representation Hypothesis~\cite{park2024linear}, our claim is that there exists a dominant low-dimensional bias direction in the final hidden-state space, which is well approximated by the vocabulary mean and which is consistently aligned with contextualized mask states across decoding steps.
A detailed comparison against alternative prior directions, including frequency-weighted lexical priors and random directions, and the stability of this alignment across generation steps, are provided in~\cref{sec:app_prior_justify}.

\begin{figure*}[!t]
\includegraphics[width=\textwidth]
{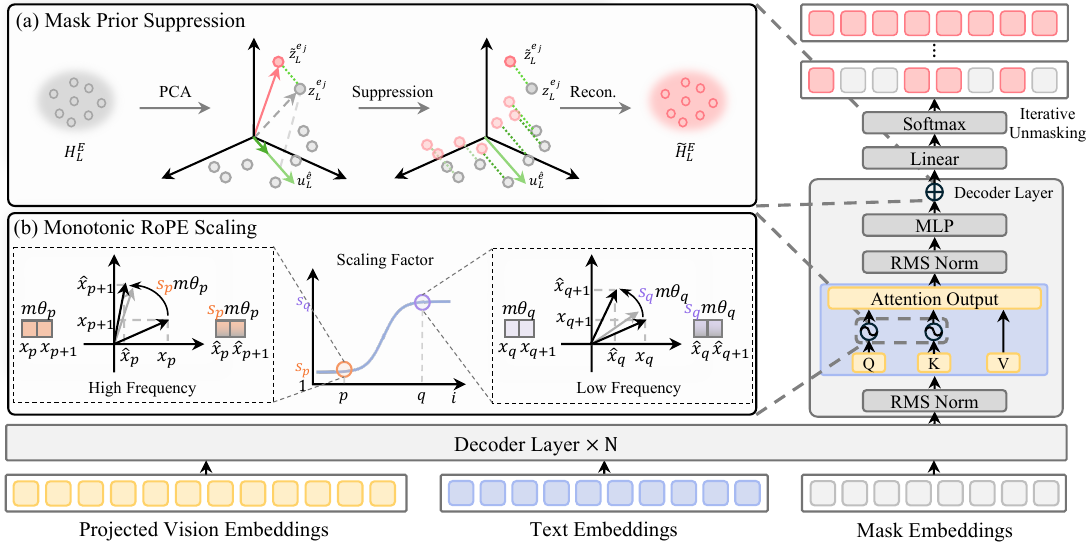}
\caption{
\textbf{Overview of the proposed model.}
(a) \emph{Mask prior suppression.} The final hidden state $h_L^{e_j}$ is decomposed along the prior direction $u^{\hat{e}}_L$, and prior components are adaptively suppressed based on cosine similarity.
(b) \emph{Monotonic RoPE scaling.} 
Low-frequency RoPE components, which govern long-range positional interactions, are scaled more strongly than high-frequency components to preserve attention to distant visual tokens, where $m$ denotes the token position.
}
\vspace{-10pt}
\label{fig:pipeline}
\end{figure*}

\vspace{-5pt}
\subsection{Positional Attention Collapse}
\vspace{-5pt}
\textbf{Positional bias under iterative unmasking.}
Beyond token repetition, LDVLMs often exhibit degraded visual grounding.
This degradation arises from a structural misalignment between the locality bias induced by RoPE-based attention and the iterative unmasking process.
Motivated by this observation, we analyze how RoPE interacts with the generation dynamics of LDVLMs.
While the long-range decay property of RoPE~\cite{roformer} facilitates coherence in autoregressive language modeling, it introduces a critical bottleneck for VLMs, which require more global access to visual information~\cite{longrope2024,wang2025circle,wei2025videorope,li2025hope}.
This limitation is further exacerbated in LDVLMs, where generation proceeds via iterative unmasking under bidirectional attention.
Generation tokens originate from semantically uninformative mask tokens.
During iterative unmasking, close positional proximity leads to disproportionately high attention.

\vspace{-3pt}
\textbf{Attention collapse toward mask tokens.}
To investigate this behavior, we analyze attention distributions during generation.
As illustrated in~\cref{fig:freq}(a), generation tokens allocate substantial attention weights to mask tokens whose semantic representations are not fully formed.
Moreover,~\cref{fig:freq}(b) shows that mask tokens receive a total amount of attention comparable to that of visual tokens across generation steps.
As a result, attention to distant visual information is suppressed, leading to degraded visual grounding.

\vspace{-3pt}
\textbf{Frequency-wise decomposition of attention.}
To further examine this structural limitation, we decompose RoPE-based attention into frequency components and examine their respective contributions to attention allocation.
RoPE implicitly orders embedding dimensions by frequency, where each pair of dimensions is associated with a predefined frequency $\theta_i$:
lower indices correspond to higher-frequency components, while higher indices capture lower-frequency components.
As shown in~\cref{fig:freq}(c), attention from generation tokens to mask tokens (Gen$\rightarrow$Mask) is dominated by high-frequency components.
This dominance induces strong local attention concentration among generation tokens.
In contrast, attention from generation tokens to visual tokens (Gen$\rightarrow$Vis) increasingly relies on low-frequency components as relative distance grows, reflecting their role in supporting long-range interactions.
However, despite this shift, overall attention still decays with distance, leaving distant visual tokens weakly attended.
Overall, these results indicate that while low-frequency components partially support long-range attention, RoPE's frequency-dependent behavior ultimately biases attention toward nearby generation tokens, limiting effective visual grounding.
\vspace{-7pt}
\section{Method}
Motivated by the analysis in~\cref{sec:analysis}, we propose two training-free approaches to mitigate the mask prior drift and attention collapse identified in LDVLMs.
As illustrated in ~\cref{fig:pipeline}, we first introduce a \textit{Mask Prior Suppression} mechanism to debias the inherent prior in mask embedding hidden states.
Subsequently, we employ \textit{Monotonic RoPE Scaling} to alleviate visual grounding degradation arising from vision-language misalignment.
\vspace{-5pt}

\subsection{Mask Prior Suppression}
\vspace{-2pt}
Our analysis reveals that mask token representations tend to drift toward the vocabulary mean in the final hidden states.
To address this issue, we apply a prior suppression mechanism to the final hidden states.
Let $E=(e_1,\dots,e_J)$ denote an input embedding sequence of length $J$, where $e_j$ is the embedding of the $j$-th token.
Let $H_l^{E}=f_{l}(E)=(h^{e_1}_l,\dots,h^{e_J}_l)$ denote the corresponding hidden states at the $l$-th transformer layer, where $h^{e_j}_l$ is the hidden state corresponding to the $j$-th embedding at layer $l$.
We further denote by $\mathcal{S}_{\mathcal{M}} \subseteq \{1,\dots,J\}$ the set of token positions occupied by mask tokens, i.e., $j \in \mathcal{S}_{\mathcal{M}}$ iff $e_j = e_{\mathcal{M}}$.

\vspace{-3pt}
\textbf{Prior subspace construction.}
To capture the intrinsic prior geometry induced by an uncontextualized masked token,
we approximate the mask token prior using the mean vocabulary embedding $\hat{e}$, which represents a context-agnostic average over token semantics.
This choice is motivated by the Linear Representation Hypothesis~\cite{park2024linear}, which suggests that even in nonlinear models, meaningful concepts can emerge as approximately linear directions in representation space.
We empirically verify in~\cref{sec:app_prior_justify} that the vocabulary mean is highly aligned with both a frequency-weighted lexical prior and contextualized mask-token states, while a random direction is not, supporting its use as a stable, parameter-free proxy for the dominant collapse direction.
We thus forward $\hat{e}$ through the transformer layers to obtain a set of layer-wise mean representations $\{h_l^{\hat{e}}\}_{l=1}^{L}$.
We then compute their mean $\mu=\frac{1}{L}\sum_{l=1}^{L}h_l^{\hat{e}}$ and perform PCA on the centered representations,
yielding an orthonormal basis $U\in\mathbb{R}^{d\times k}$ that spans the mean prior subspace, where $k$ denotes the number of principal components.

\vspace{-3pt}
\textbf{Projection and similarity estimation.}
For each token $e_j$, we define its deviation in the final layer as $\delta^{e_j}_L = h^{e_j}_L - \mu$. This deviation is projected onto the prior subspace: $z^{e_j}_L = U^\top \delta^{e_j}_L \in \mathbb{R}^k$. 
Similarly, the projected prior direction is denoted as $z^{\hat{e}}_L=U^\top(h^{\hat{e}}_L-\mu)$  with its normalized unit vector
$u^{\hat{e}}_L=\frac{z^{\hat{e}}_L}{\|z^{\hat{e}}_L\|_2}$.
We quantify the alignment between a token and the mask prior using cosine similarity:
\[
c^{e_j} = \frac{\langle z^{e_j}_L, u^{\hat{e}}_L \rangle}{\|z^{e_j}_L\|_2}.
\]

\vspace{-3pt}
\textbf{Adaptive prior suppression.}
We attenuate components that are aligned with the prior direction in proportion to this similarity.
Let $\alpha^{e_j} = \lambda \max(0, c^{e_j})$ be the suppression strength, where $\lambda$ is a positive hyperparameter. 
The updated projection is computed as:
\[
\tilde{z}^{e_j}_L = z^{e_j}_L - \alpha^{e_j} \langle z^{e_j}_L, u^{\hat{e}}_L \rangle u^{\hat{e}}_L.
\]
The modified projection $\tilde{z}^{e_j}_L$ is then reconstructed in the original feature space as $\tilde{h}^{e_j}_L = U \tilde{z}^{e_j}_L+\mu$,
which selectively removes only the component associated with the mask prior.
This operation is applied to all masked positions, i.e., $\forall j\in\mathcal{S}_{\mathcal{M}}$, effectively preserving the semantic integrity of unmasked tokens while preventing representational collapse in the generated sequence.
Therefore, by applying this operation to all masked positions, we obtain the updated final hidden states $\tilde{H}_L^{E}$ from the original $H_L^{E}$, as illustrated in~\cref{fig:pipeline}(a).

\textbf{Remark on intervention scale.}
Mask Prior Suppression modifies only $k=3$ out of $d=4096$ dimensions ($\sim$0.07\%) at the final layer, comparable in scale to representation-level interventions used in autoregressive LLMs without retraining~\cite{li2023iti,wang2025sadi}.
This small footprint, combined with the consistent gains we observe across multiple LDVLM architectures and benchmarks (\cref{tab:main_results,tab:app_unified}), suggests that the modification does not broadly disrupt the learned diffusion dynamics.

\vspace{-7pt}
\subsection{Monotonic RoPE Scaling}
\vspace{-3pt}
We introduce a frequency-aware adjustment to RoPE that applies monotonic scaling across frequency components.
Rather than uniformly scaling all RoPE frequencies, our method progressively modulates the scaling strength according to the frequency index.
\begin{table*}[!t]
\centering
\small
\caption{Comparison of LDVLMs on general benchmarks, visual grounding, and long-form generation tasks.
\textbf{Bold}: best, \underline{underline}: second best. AR: Autoregressive models, Diff.: LDVLMs.
}
\resizebox{\textwidth}{!}{%
\begin{tabular}{l c ccccccccc}
\toprule
\multirow{2}{*}{Model}
& \multirow{2}{*}{Type}
& \multicolumn{3}{c}{General}
& \multicolumn{3}{c}{Visual Grounding}
& \multicolumn{3}{c}{Long-form Generation} \\
\cmidrule(lr){3-5} \cmidrule(lr){6-8} \cmidrule(lr){9-11}

& 
& MME (sum)$\uparrow$
& MMBench$\uparrow$
& MMMU$\uparrow$
& RefCOCOg$\uparrow$
& Ferret$\uparrow$
& GQA$\uparrow$
& LLaVA-Bench$\uparrow$
& DetailCaps$\uparrow$
& MIA$\uparrow$\\
\midrule

LLaVA-OV
& AR
& 1998 & 80.8 & 48.8
& 56.9 & 36.0 & 72.2
& 83.8 & 60.6 & 76.1 \\

Qwen2.5
& AR
& 2448 & 83.5 & 58.6
& 16.6 & 53.8 & 70.6
& 96.6 & 63.3 & 84.6 \\

InternVL3
& AR
& 2415 & 83.4 & 53.6
& 67.1 & 59.0 & 71.2
& 82.3 & 64.4 & 82.3 \\
\midrule

LLaVA-1.6
& AR
& 1842 & 68.1 & 42.6
& 28.1 & 60.3 & \textbf{75.8}
& \textbf{75.3} & \underline{60.6} & \textbf{76.1} \\

LLaDA-V
& Diff.
& \underline{1998} & \underline{82.9} & \underline{48.6}
& \underline{64.8} & \underline{60.4} & \underline{61.6}
& {61.3} & {59.8}& {66.1} \\

\rowcolor{gray!30}
{+ Ours}
& Diff.
& \textbf{2003} &  \textbf{83.3} &  \textbf{49.3}
&  \textbf{65.0} &  \textbf{62.9} &  \underline{61.6}
&  \underline{64.1} & \textbf{63.6} & \underline{67.0} \\

LaViDa
& Diff.
& 1682 & 71.7 & 43.2
& 36.9 & 25.9 & 59.4
& 39.5 & 8.3 & 49.4 \\

\rowcolor{gray!30}
{+ Ours}
& Diff.
& {1705} & {72.0} & {43.7}
& {44.0} & {35.7} & {60.2}
& {46.5} & {56.1} & {57.3} \\
\bottomrule
\end{tabular}
}
\label{tab:main_results}
\vspace{-7pt}
\end{table*}

\begin{figure*}[!t]
\includegraphics[width=\textwidth]
{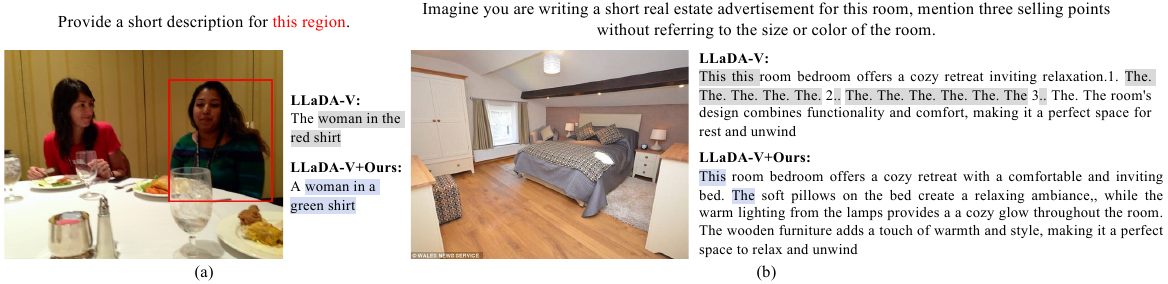}
\caption{
\textbf{Qualitative comparison on visual grounding and long-form generation.}
(a) RefCOCOg results.
A \red{red} box indicates the target region.
The baseline model LLaDA-V produces descriptions referring to an incorrect object shown in \gray{gray}, whereas our method correctly grounds the description to the target location and achieves more accurate visual grounding shown in \blue{blue}.
(b) MIA results.
The baseline model exhibits repetitive descriptions shown in \gray{gray}.
Our method shown in \blue{blue} effectively mitigates repetition and generates more coherent and informative long-form descriptions.
}
\vspace{-10pt}
\label{fig:qual}
\end{figure*}

\vspace{-3pt}
\textbf{RoPE scaling function.}
We first normalize the frequency index $i\in\{0,\ldots,d/2-1\}$ as $\tau_i = {i}/{(d/2 - 1)}.$
Recall that $\theta_i = 10000^{-2i/d}$, so larger $i$ corresponds to a lower RoPE frequency.
We define a sigmoid gate
\[
\gamma_i = \sigma\!\bigl(\eta(\tau_i - \tau_0)\bigr),
\qquad 
\sigma(\xi)=\frac{1}{1+\exp(-\xi)},
\]
where $\tau_0$ specifies the frequency boundary at which scaling becomes active and $\eta>0$ controls the sharpness of the transition.
The  scaling factor is then given by
\[
s_i = 1 + \beta\,\gamma_i,
\]
where $\beta\ge 0$ controls the maximum scaling.

\vspace{-3pt}
\textbf{Frequency-aware modulation.}
This formulation induces a smooth and monotonic scaling schedule along the frequency axis.
High-frequency components are assigned scaling factors close to $1$, while lower-frequency components are progressively amplified, with the overall scaling smoothly bounded within $(1,\,1+\beta)$.
We apply frequency-aware scaling to the RoPE rotation angle as
\[
m\theta_i \;\leftarrow\; s_i\, m\theta_i.
\]
As a result, high-frequency positional information is largely preserved, whereas lower-frequency components receive stronger scaling, which prevents attention collapse at large positional distances, as shown in~\cref{fig:pipeline}(b).
\section{Experiment}
\vspace{-3pt}
\textbf{Implementation details.}
We implement our proposed method on top of LLaDA-V~\cite{you2025llada} and LaViDa~\cite{li2025LaViDa},
and observe consistent trends across both backbones.
Additional analyses for each model are provided in~\cref{sec:app_analysis2}.
For the LLaDA-V-based implementation, we set the hyperparameters to $\lambda=0.1$, $\beta=0.01$, and $\eta=8.0$, while for LaViDa, we use $\lambda=0.3$, $\beta=0.01$, and $\eta=12.0$.
We fix $\tau_0=0.6$ and $k=3$ throughout all experiments.
All hyperparameters are fixed across tasks and datasets for each backbone.
To ensure a fair comparison with existing works, all evaluations are conducted using the LMMs-Eval~\cite{llms-eval} framework. 
Further hyperparameter sensitivity experiments are presented in~\cref{sec:app_hyper}, and detailed evaluation setups are described in~\cref{sec:app_eval}.

\vspace{-3pt}
\textbf{Datasets.}
To comprehensively evaluate the general reasoning, visual grounding, and long-form generation capabilities of our model, we conduct experiments across nine widely recognized benchmarks. 
For general-purpose evaluation, we adopt MME~\cite{mme}, MMBench~\cite{MMBench}, and MMMU~\cite{mmmu}. 
To assess visual grounding, we utilize RefCOCOg~\cite{refcocog}, Ferret~\cite{ferret}, and GQA~\cite{gqa}. Finally, for long-form generation, we use LLaVA-Bench~\cite{liu2023visual}, DetailCaps~\cite{detailcaps}, and MIA~\cite{mia}.
For computational efficiency, we adopt GQA Lite for GQA, and randomly sampled 100 instances for both DetailCaps and Ferret.
Further details are provided in~\cref{sec:benchmark}.
\vspace{-5pt}

\subsection{Main Results}
\label{sec:main_results}

\vspace{-2pt}
\textbf{Overall performance.}
\cref{tab:main_results} summarizes the performance of our method across general reasoning, visual grounding, and long-form generation tasks.
Our approach achieves comparable or superior performance across the evaluated benchmarks without requiring any additional training.
We compare to LLaVA-1.6-7B~\cite{llava16}, which is trained with a comparable data scale and parameter count, and further include baselines of similar model size trained on larger datasets, including LLaVA-One-Vision-7B~\cite{llavaonevision}, Qwen2.5-VL-7B~\cite{qwen2}, and InternVL-3-8B~\cite{invternvl3}.
Additional details on the compared models are provided in~\cref{sec:app_comparison}.
For LLaDA-V, our method demonstrates clear improvements on visual grounding benchmarks such as RefCOCOg and Ferret, as well as on the long-form generation benchmark LLaVA-Bench and DetailCaps, while maintaining competitive performance on general reasoning tasks.
A similar trend is observed for LaViDa, where grounding accuracy improves substantially alongside consistent gains in long-form generation quality.
The degraded quality of the LaViDa baseline is further investigated through a detailed performance analysis in~\cref{sec:app_analysis2}.

\begin{figure}[t]
\includegraphics[width=\linewidth]
{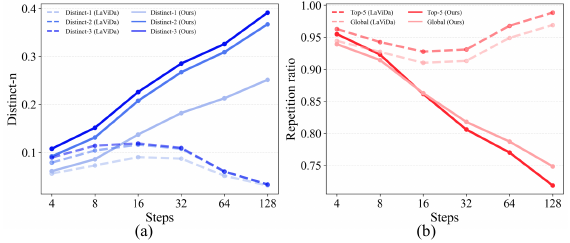}
\caption{
\textbf{Results of LaViDa on DetailCaps with varying generation steps.}
Dashed lines: LaViDa, solid lines: Ours.
(a) Distinct-$n$ scores of our method exhibit an increasing trend across the evaluated settings and remain consistently higher than those of the baseline.
(b) The repetition ratio under our method shows a decreasing trend across the evaluated settings and remains consistently lower than that of the baseline.
}
\vspace{-10pt}
\label{fig:quant}
\end{figure}

\vspace{-3pt}
\textbf{Robustness across generation steps.}
As illustrated in~\cref{fig:quant}, our approach consistently produces coherent and diverse outputs across different generation settings.
Due to the mask token prior, baseline LDVLMs tend to repeatedly generate mean embedding prior tokens, which often manifests as repeated generation of specific words.
This behavior leads to premature termination through excessive emission of the \texttt{|eot|} token or results in a degraded lexical diversity in the generated descriptions.
In contrast, our method effectively suppresses the mask token prior, mitigating repetitive generation and enabling more informative and diverse long-form outputs.

\vspace{-3pt}
\textbf{Qualitative results.}
Qualitative examples in~\cref{fig:qual} illustrate that our method mitigates failures of LDVLMs.
Additional results on unified vision--language architectures and qualitative examples are provided in~\cref{sec:app_unified,sec:app_qual}.

\begin{table}[t]
\centering
\caption{Ablation study on visual grounding and long-form generation benchmarks across LLaDA-V and LaViDa.
\textbf{Bold}: best, \underline{underline}: second best; 
MPS: Mask Prior Suppression; MRS: Monotonic RoPE Scaling.}
\resizebox{0.99\linewidth}{!}{%
\begin{tabular}{lcccccc}
\toprule
\multirow{2}{*}{Model}
& \multicolumn{2}{c}{Method}
& \multicolumn{2}{c}{Visual Grounding}
& \multicolumn{2}{c}{Long-form Generation} \\
\cmidrule(lr){2-3}
\cmidrule(lr){4-5}
\cmidrule(lr){6-7}

&
MPS
& MRS
& RefCOCOg
& Ferret
& LLaVA-Bench
& DetailCaps \\
\midrule

\multirow{4}{*}{LLaDA-V}
& 
& 
& \underline{64.8}
& 60.4
& 61.3
& 59.8 \\

& \checkmark
& 
& \underline{64.8}
& 60.2
& 61.7
& \underline{60.0} \\

& 
& \checkmark
& \underline{64.8}
& \underline{60.6}
& \underline{63.9}
& \underline{60.0} \\

& \checkmark
& \checkmark
& \textbf{65.0}
& \textbf{62.9}
& \textbf{64.1}
& \textbf{63.6} \\
\midrule

\multirow{4}{*}{LaViDa}
& 
& 
& 36.9
& 25.9
& 39.5
& \underline{8.3} \\

& \checkmark
& 
& \textbf{44.0}
& \underline{35.4}
& \underline{42.3}
& \textbf{56.1} \\

& 
& \checkmark
& \underline{39.5}
& 29.8
& 41.0
& \underline{8.3} \\

& \checkmark
& \checkmark
& \textbf{44.0}
& \textbf{35.7}
& \textbf{46.5}
& \textbf{56.1} \\

\bottomrule
\end{tabular}
}
\label{tab:ablation}
\vspace{-10pt}
\end{table}

\subsection{Ablation Study}
\textbf{Ablation on MPS and MRS.}
We conduct an ablation study to analyze the effects of Mask Prior Suppression (MPS)
and Monotonic RoPE Scaling (MRS) on long-form generation and visual grounding
tasks, as summarized in~\cref{tab:ablation}.
Overall, the full model exhibits the most stable performance across benchmarks,
indicating the benefit of addressing both mask prior drift and positional
attention bias.
On LLaDA-V, removing either component results in clear performance drops across
tasks, suggesting that MPS and MRS play complementary roles.
In particular, removing MPS tends to have a larger impact on long-form
generation benchmarks such as LLaVA-Bench and DetailCaps, while removing MRS
more noticeably affects visual grounding performance on RefCOCOg and Ferret.
For LaViDa, individual components can be competitive on certain benchmarks. 
For instance, removing MRS yields results comparable to those on RefCOCOg and DetailCaps.
Nevertheless, the full model provides the best overall performance across tasks,
supporting the importance of combining both mechanisms for robust and balanced
multimodal generation.

\textbf{Choice of mask prior direction.}
We analyze the choice of mask prior direction while keeping Monotonic RoPE Scaling fixed, as shown in~\cref{tab:emb}.
The vocabulary mean direction consistently yields the largest and most stable gains across both visual grounding and long-form generation benchmarks.
This indicates that, beyond the overall performance improvement introduced by RoPE scaling, effective mask prior suppression critically depends on principled prior direction selection rather than subtracting an arbitrary component.
A comparison against a decoding-level baseline (DPad-style suffix dropping) is provided in~\cref{sec:app_dpad}, and further results in~\cref{sec:app_result}.

\textbf{Comparison with RoPE scaling baselines.}
We further verify that conventional RoPE rescaling does not address the locality bias under iterative unmasking.
\cref{tab:rope_baseline} compares NTK-aware scaling~\cite{ntk,liu2024scaling} and YaRN~\cite{yarn} as drop-in replacements for our Monotonic RoPE Scaling.
Both baselines, designed for long-context extrapolation, often degrade visual grounding, whereas our monotonic low-frequency amplification yields consistent gains across both backbones.

\begin{table}[t]
\centering
\small
\caption{Comparison with representative RoPE scaling baselines.}
\setlength{\tabcolsep}{4pt}
\resizebox{0.99\linewidth}{!}{%
\begin{tabular}{lccc|ccc}
\toprule
& \multicolumn{3}{c|}{LLaDA-V} & \multicolumn{3}{c}{LaViDa} \\
\cmidrule(lr){2-4} \cmidrule(lr){5-7}
Method & RefCOCOg & Ferret & DetailCaps & RefCOCOg & Ferret & DetailCaps \\
\midrule
Baseline & 64.8 & 60.4 & 59.8 & 36.9 & 25.9 & 8.3 \\
+ NTK    & 61.9 & 59.6 & 60.9 & 39.6 & 30.9 & 7.9 \\
+ YaRN   & 59.3 & 56.0 & 59.1 & 39.6 & 30.7 & 7.9 \\
\rowcolor{gray!30}
+ Ours   & \textbf{65.0} & \textbf{62.9} & \textbf{63.6} & \textbf{44.0} & \textbf{35.7} & \textbf{56.1} \\
\bottomrule
\end{tabular}}
\label{tab:rope_baseline}
\vspace{-10pt}
\end{table}

\subsection{Result Analysis}
\label{sec:result_analysis}
\begin{figure}[t]
\includegraphics[width=\linewidth]
{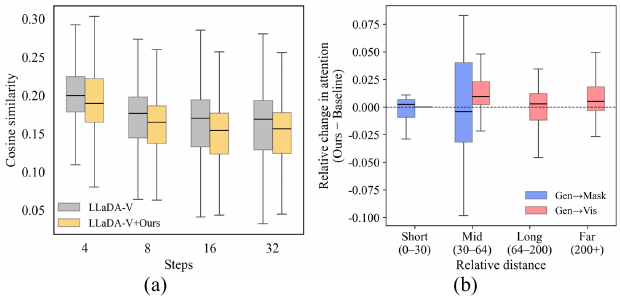}
\caption{
\textbf{Visualization of result analysis.}
(a) Box plot of cosine similarity between contextualized mask tokens and the vocabulary mean, showing consistent reduction across generation steps.
(b) Relative change in attention with respect to relative distance, where attention to distant visual tokens increases compared to the baseline, while attention to mask tokens is preserved or reduced.
}   
\label{fig:result_analysis}
\vspace{-10pt}
\end{figure}

\vspace{-2pt}
\textbf{Effect of Mask Prior Suppression.}
To quantify mask prior drift during decoding, we measure the cosine similarity between the contextualized last layer hidden states of mask tokens and the vocabulary mean prior.
High similarity reflects increasing alignment with a shared prior direction, leading to reduced semantic diversity and repetitive generation.
As illustrated in~\cref{fig:result_analysis}(a), our method consistently reduces this cosine similarity across generation steps.
This observation indicates that suppressing a low-dimensional prior subspace is sufficient to prevent mask prior drift during decoding, without modifying the full representational space.
Consequently, the model produces less repetitive and more semantically diverse generations.
Moreover, since mask prior suppression operates on a low-dimensional subspace, it introduces minimal inference overhead, as confirmed by the latency analysis in~\cref{sec:latency}.

\begin{table}[t]
\centering
\small
\caption{Ablation study on different mask prior directions.
We compare vocabulary-mean prior (ours), top-$k$ token-based priors,
and random directions on visual grounding and long-form generation benchmarks.}
\resizebox{0.99\linewidth}{!}{%
\begin{tabular}{llcccc}
\toprule
\multirow{2}{*}{Model}
& \multirow{2}{*}{Type}
& \multicolumn{2}{c}{Visual Grounding}
& \multicolumn{2}{c}{Long-form Generation} \\
\cmidrule(lr){3-4}
\cmidrule(lr){5-6}

& 
& RefCOCOg
& Ferret
& LLaVA- Bench
& MIA \\
\midrule

\multirow{5}{*}{LLaDA-V}

& Top-1
& 64.4 & 61.9 & 60.8 & 66.4 \\
& Top-3
& 65.0 & 62.4 & 65.3 & 65.4 \\
& Random1
& 65.0 & 61.8 & 62.0 & 66.5 \\
& Random2
& 64.7 & 61.0 & 62.9 & 66.4 \\
& \cellcolor{gray!30}\textbf{Ours}
& \cellcolor{gray!30}65.0
& \cellcolor{gray!30}62.9
& \cellcolor{gray!30}64.1
& \cellcolor{gray!30}67.0 \\

\midrule

\multirow{5}{*}{LaViDa}

& Top-1
& 44.0 & 34.8 & 44.5 & 57.1 \\
& Top-3
& 44.0 & 35.3 & 44.6 & 56.0 \\
& Random1
& 44.0 & 34.8 & 45.1 & 57.0 \\
& Random2
& 44.0 & 34.6 & 41.6 & 56.9 \\
& \cellcolor{gray!30}\textbf{Ours}
& \cellcolor{gray!30}44.0
& \cellcolor{gray!30}35.7
& \cellcolor{gray!30}46.5
& \cellcolor{gray!30}57.3 \\

\bottomrule
\end{tabular}}
\label{tab:emb}
\end{table}

\vspace{-2pt}
\textbf{Effect of Monotonic RoPE Scaling.}
To analyze how monotonic RoPE scaling affects positional attention, we examine changes in attention allocation with respect to relative token distance.
Specifically, we distinguish attention to nearby mask tokens from attention to distant visual tokens, which is critical for visual grounding in LDVLMs.
As shown in~\cref{fig:result_analysis}(b), monotonic RoPE scaling gradually amplifies low-frequency components associated with long-range positional relations, leading to a redistribution of attention.
Attention to nearby, semantically neutral mask tokens is reduced, while attention is reallocated toward distant visual tokens.
By alleviating the misalignment between RoPE's locality bias and the bidirectional unmasking mechanism of LDVLMs, this adjustment preserves global access to visual representations and stabilizes visual grounding.

\textbf{Effect on hallucination.}
To examine whether the improvements in visual grounding translate into reduced hallucination, rather than mere suppression of generation,
we evaluate our method on two hallucination-focused benchmarks: CHAIR~\cite{rohrbach2018chair} on the COCO validation split (500 images), and AMBER-G~\cite{wang2023amber} on its full discriminative split (1{,}004 images).
As shown in~\cref{tab:hallucination}, our method consistently reduces both CHAIR$_s$ and CHAIR$_i$ on LLaDA-V, while AMBER-G shows lower CHAIR and HAL with coverage preserved.
This indicates that the gains stem from improved visual grounding rather than degenerate or overly conservative generation.
These results are consistent with prior findings that strengthening attention to visual tokens reduces hallucination~\cite{huang2024opera,arif2025paint,favero2024m3id}; \cref{fig:result_analysis}(b) confirms that our method increases attention from generation tokens toward distant visual tokens.

\begin{table}[t]
\centering
\small
\caption{\textbf{Hallucination evaluation on CHAIR and AMBER-G.}
Lower is better for CHAIR$_s$, CHAIR$_i$, AMBER CHAIR, and HAL; higher is better for Cover.
}
\setlength{\tabcolsep}{4pt}
\begin{tabular}{lccccc}
\toprule
& \multicolumn{2}{c}{CHAIR (COCO)} & \multicolumn{3}{c}{AMBER-G} \\
\cmidrule(lr){2-3} \cmidrule(lr){4-6}
Method & CHAIR$_s\downarrow$ & CHAIR$_i\downarrow$ & CHAIR$\downarrow$ & Cover$\uparrow$ & HAL$\downarrow$ \\
\midrule
LLaDA-V & 29.4 & 9.5 & 7.2 & 61.6 & 41.1 \\
\rowcolor{gray!30}
+ Ours  & \textbf{27.0} & \textbf{8.3} & \textbf{6.9} & 61.6 & \textbf{40.6} \\
\bottomrule
\end{tabular}
\label{tab:hallucination}
\vspace{-10pt}
\end{table}

\textbf{Effect on visual spatial structure.}
A natural concern is that frequency rescaling may distort the 2D visual spatial structure when applied uniformly across token segments.
We verify this is not the case: applying MRS only to visual tokens already improves grounding on both backbones (e.g., $36.9 \rightarrow 43.8$ on RefCOCOg for LaViDa), indicating that monotonic low-frequency amplification preserves 2D positional reasoning rather than damaging it.
The full segment-wise comparison (visual-only / textual-only / generation-only / all) is provided in~\cref{sec:app_mrs_segment}.
\section{Conclusion}
In this work, we analyze fundamental challenges of LDVLMs, including repetitive generation and degraded visual grounding.
We identify mask representation drift and positional attention collapse as the primary causes of these issues.
To mitigate them, we propose two training-free inference-time techniques: Mask Prior Suppression and Monotonic RoPE Scaling.
Our methods integrate seamlessly into existing LDVLMs and consistently improve visual grounding and long-form generation across diverse benchmarks, without sacrificing general reasoning performance.
As an inference-time approach, the effectiveness of our method depends on the quality of the underlying pretrained representations.
Our experiments are conducted under the standard full-suffix iterative unmasking setup adopted by current LDVLMs;
preliminary results in~\cref{sec:app_dpad} suggest that our methods remain effective when suffix redundancy is partially reduced (e.g., DPad-style decoding),
but extending the analysis to alternative decoding paradigms such as semi-autoregressive or block-wise diffusion variants requires additional multimodal adaptation and is left to future work.
We hope this work encourages further research toward more stable and interpretable LDVLMs.


\clearpage
\section*{Impact Statement}
This work focuses on improving the stability and reliability of LDVLMs through analysis and inference-time techniques.
However, as with other advances in multimodal generation, these techniques could also be applied to systems that generate misleading or harmful content.
We emphasize that our methods do not introduce new capabilities beyond those of existing models, and that responsible deployment should follow established safety and ethical guidelines.


\bibliography{main}
\bibliographystyle{icml2026}

\newpage
\appendix
\onecolumn

\section{Benchmark Details}
\label{sec:benchmark}

\subsection{General}
\myparagraph{MME.}~\cite{mme}
The MME benchmark is designed to evaluate both perceptual and cognitive abilities of multimodal models.
It consists of 14 subtasks, each formulated as a binary (yes/no) question.
The perceptual evaluation includes tasks such as object existence and object counting, while the cognitive evaluation covers tasks including commonsense reasoning and text translation.

\myparagraph{MMBench.}~\cite{MMBench}
The MMBench evaluates the multidimensional capabilities of multimodal models using 3,217 multiple-choice questions organized into a three-level hierarchy.
It covers perception and reasoning at the first level, six specialized capability groups (e.g., attribute reasoning and coarse perception) at the second level, and 20 fine-grained abilities (e.g., physical properties and image style) at the final level.

\myparagraph{MMMU.}~\cite{mmmu}
The MMMU is a benchmark designed to evaluate advanced academic-level knowledge and reasoning capabilities of multimodal models.
It consists of six subject categories: Art \& Design, Business, Science, Health \& Medicine, Humanities \& Social Science, and Technology \& Engineering.
The benchmark includes diverse visual modalities beyond natural images, such as charts, musical scores, and chemical structure diagrams.
Due to its inclusion of complex and high-level tasks, MMMU is regarded as one of the most challenging benchmarks for MLLMs.

\subsection{Visual Grounding}

\myparagraph{RefCOCOg.}~\cite{refcocog}
RefCOCOg is a benchmark designed to evaluate a model's ability to comprehend referring expressions and segment specific objects. It is composed of samples that involve long descriptions and demand complex reasoning, presenting a high level of difficulty.

\myparagraph{Ferret.}~\cite{ferret}
Ferret evaluates referring and grounding capabilities, which are divided into description, referring reasoning, and grounding in conversational settings.
The dataset is constructed by sampling 40 images from the COCO validation set~\cite{coco} and generating corresponding reference responses using GPT-4.

\myparagraph{GQA.}~\cite{gqa}
The GQA is a benchmark for real-world visual question answering (VQA).
In addition to images and questions, it incorporates scene graphs that encode object attributes, spatial positions, and relationships between objects.
The dataset balances answer distributions to reduce language priors.
Through this benchmark, models can be evaluated on their ability to understand complex relational structures within images.

\subsection{Long-form Generation}
\myparagraph{LLaVA Bench (In-the-Wild).}~\cite{liu2023visual}
The LLaVA-Bench (In-the-Wild) benchmark evaluates a model's creativity and comprehension through open-ended responses, focusing on challenging tasks and generalization to unseen domains.
The benchmark evaluates performance on challenging tasks and generalization to unseen domains.
It consists of 60 questions associated with 24 images covering diverse real-world scenes.

\myparagraph{DetailCaps.}~\cite{detailcaps}
DetailCaps is designed to evaluate the performance of models in image captioning tasks. Its evaluation datasets are constructed through rigorous annotation by human experts and advanced MLLMs such as GPT-4V. It is useful in mitigating hallucinations and enabling analysis of a more diverse range of visual elements. 

\myparagraph{MIA.}~\cite{mia}
The MIA benchmark evaluates whether a model correctly follows the instructions given.
It consists of 400 image–prompt pairs, where each sample includes specific constraints, such as sentence length and output format.
GPT-4o is used as an automatic judge to assess not only answer correctness but also instruction adherence.

\subsection{Hallucination}
\myparagraph{CHAIR.}~\cite{rohrbach2018chair}
CHAIR (Caption Hallucination Assessment with Image Relevance) measures object hallucination in image captioning by comparing the set of objects mentioned in a generated caption against the ground-truth object set defined by COCO annotations.
We report two standard variants: CHAIR$_s$ (sentence-level: fraction of sentences containing at least one hallucinated object) and CHAIR$_i$ (instance-level: fraction of mentioned objects that are hallucinated).

\myparagraph{AMBER-G.}~\cite{wang2023amber}
AMBER is an LLM-free hallucination benchmark covering both generative (AMBER-G) and discriminative settings.
We use the generative split, which evaluates open-ended captions against a curated object inventory and reports CHAIR (object hallucination), Cover (object recall), and HAL (hallucination rate over generated samples).

\section{Evaluation Setup.}
\label{sec:app_eval}

We conduct all evaluations on LLaDA-V and LaViDa using the LMMs-Eval framework~\cite{llms-eval}.
For all benchmarks, we adopt the default prompts provided by the framework.
The evaluation split and the generation length $L$ for each dataset
are reported in \cref{tab:app_eval}.
For autoregressive baselines, including LLaVA-One-Vision-7B, Qwen2.5-VL-7B,
InternVL3-8B, and LLaVA-1.6, we use the default evaluation setups
provided by the same framework.
To ensure a rigorous and fair comparison, we evaluate models under identical random seeds whenever reported results are unavailable. Notably, for LaViDa, we conduct a re-evaluation using the same evaluation setups as LLaDA-V, thereby facilitating a strictly controlled comparison.
\begin{table}[t]
\centering
\caption{\textbf{Evaluation Setup.} Evaluation splits, inference steps,
and generation length $L$ for each benchmark.}
\small
\begin{tabular}{lccc|lccc|lccc}
\toprule
Dataset & Split & Steps & $L$
& Dataset & Split & Steps & $L$
& Dataset & Split & Steps & $L$ \\
\midrule
MME      & test   & 2   & 2
& Ferret & test   & 48  & 96
& DetailCaps & test & 128 & 256 \\

MMBench  & en-dev & 2   & 2
& GQA    & lite   & 2   & 2
& LLaVA-Bench & train & 64 & 128 \\

MMMU     & val    & 2   & 2
& RefCOCOg & val & 4   & 8
& MIA    & test   & 32  & 64 \\
\bottomrule
\end{tabular}
\label{tab:app_eval}
\end{table}

\subsection{Details of Comparison Models}
\label{sec:app_comparison}
As shown in~\cref{tab:comparison_models}, we summarize the parameter counts and training data scale of the compared models.

\begin{table}[t]
\centering
\small
\caption{\textbf{Comparison of training data scale and model size across LDVLMs and baselines.}
Dashes (-) indicate results not reported.
L-OV; LLaVA-OneVision-7B; Qwen2.5: Qwen2.5-VL-7B; Intern3: Intern-VL3-8B
}
\begin{tabular}{l|cccccc}
\toprule
 & \textbf{LLaDA-V} & \textbf{LaViDa} & \textbf{LLaVA-1.6} & \textbf{LLaVA-OV} & \textbf{Qwen2.5} & \textbf{Intern3} \\
\midrule
\# Params & 8B & 8B & 7B & 7B & 7B & 8B \\
\# Images (Pretrain) & 0.6M & 0.6M & 0.6M & 0.6M & $>$7M & -- \\
\# Images (SFT) & 15.9M & 1.0M & 0.7M & 7.2M & $\sim$2M & 21.7M \\
\bottomrule
\end{tabular}
\label{tab:comparison_models}
\end{table}

\subsection{Inference Details.}
We conduct our experiments on a system equipped with one NVIDIA A100-SXM4-80GB GPU
and two NVIDIA RTX A6000 GPUs.
All experiments are conducted under the recommended environments
provided by the respective models to ensure fair evaluation.

\section{Justification of the Mask Prior Direction}
\label{sec:app_prior_justify}
We provide additional empirical justification for the choice of the vocabulary mean as our prior direction.
Our claim throughout the paper is empirical and limited in scope: we do not assert that the vocabulary mean fully characterizes the nonlinear dynamics of mask-token collapse, only that it captures a dominant, low-dimensional bias direction in the final hidden-state space that is consistently aligned with contextualized mask states.
Motivated by the Linear Representation Hypothesis~\cite{park2024linear}, which suggests that meaningful concepts in deep models can emerge as approximately linear directions in representation space, we adopt this simple and parameter-free proxy.

\myparagraph{Comparison against alternative prior directions.}
We compare four candidate directions in the final hidden-state space of LLaDA-V: the vocabulary-mean direction $e_{\text{uniform}}$ (ours), a frequency-weighted lexical prior $e_{\text{freq}}$ estimated from COCO train2014 caption frequencies, the contextualized mask-token direction $e_{\text{mask}}$ averaged over generation, and a random direction $e_{\text{random}}$.
\cref{tab:app_prior_justify_dirs} reports cosine similarities between these candidate directions.
The vocabulary-mean direction is highly aligned with the frequency-weighted lexical prior ($0.897$), indicating that they capture a very similar dominant lexical direction in practice.
Both $e_{\text{uniform}}$ and $e_{\text{freq}}$ are substantially more aligned with $e_{\text{mask}}$ ($0.883$, $0.735$) than a random direction ($0.427$), supporting the use of the vocabulary mean as a stable proxy for a shared collapse direction.

\begin{table}[t]
\centering
\small
\caption{\textbf{Cosine similarity between candidate prior directions in the final hidden-state space (LLaDA-V).}
$e_{\text{uniform}}$: vocabulary mean (ours); $e_{\text{freq}}$: frequency-weighted lexical prior;
$e_{\text{mask}}$: contextualized mask-token direction averaged over generation; $e_{\text{random}}$: random direction.}
\begin{tabular}{ccc}
\toprule
Direction A & Direction B & Cosine Similarity \\
\midrule
$e_{\text{uniform}}$ & $e_{\text{freq}}$   & $+0.897$ \\
$e_{\text{uniform}}$ & $e_{\text{mask}}$   & $+0.883$ \\
$e_{\text{freq}}$    & $e_{\text{mask}}$   & $+0.735$ \\
$e_{\text{random}}$  & $e_{\text{mask}}$   & $+0.427$ \\
\bottomrule
\end{tabular}
\label{tab:app_prior_justify_dirs}
\end{table}

\myparagraph{Stability across generation steps.}
We further verify that the alignment between the context-free prior direction and contextualized mask-token states is stable across decoding configurations.
\cref{tab:app_prior_justify_steps} reports the cosine similarity between the vocabulary mean and the contextualized mask-token hidden states for $T \in \{32, 16, 8, 4\}$ generation steps.
The alignment is consistently strong and varies only mildly with $T$, while a random direction yields substantially lower alignment.
This stability supports the use of a context-free prior estimate throughout iterative unmasking.

\begin{table}[t]
\centering
\small
\caption{\textbf{Stability of context-free prior alignment across generation steps.}
Cosine similarity between the vocabulary mean and contextualized mask-token hidden states for varying generation steps $T$ (LLaDA-V).}
\begin{tabular}{lcccc}
\toprule
$T$ & 32 & 16 & 8 & 4 \\
\midrule
Cosine similarity & 0.255 & 0.256 & 0.268 & 0.283 \\
\bottomrule
\end{tabular}
\label{tab:app_prior_justify_steps}
\end{table}

\myparagraph{Discussion.}
Together with~\cref{tab:emb} in the main paper, which compares the vocabulary mean against top-$k$ token-based priors and random directions in terms of downstream task performance, these results support our design choice on three grounds:
(i) the vocabulary mean and frequency-weighted lexical prior are nearly equivalent dominant directions;
(ii) both align substantially more with mask-token hidden states than random directions; and
(iii) this alignment is stable across decoding configurations.
We therefore use the vocabulary mean as a simple, principled proxy for the shared bias component, rather than as a fundamentally different prior from those propagated through the network.

\section{Limitations of Existing Rotary Scaling Methods in LDVLMs}
\label{sec:app_rope_baseline}
Prior work has explored various rotary scaling strategies to extend the context
length of autoregressive language models.
NTK-aware RoPE scaling~\cite{ntk,liu2024scaling} modifies the base frequency of rotary embeddings based on Neural Tangent Kernel analysis, stabilizing attention patterns when extending the context window beyond the training length.
Similarly, YaRN~\cite{yarn} introduces a piecewise frequency rescaling scheme that preserves high-frequency components while smoothly extrapolating to longer sequences.
Subsequent methods further refine rotary scaling to enhance extrapolation stability and efficiency in LLMs~\cite{longrope2024}.
While these approaches are effective for extending context length under causal decoding, they are primarily designed for autoregressive language modeling.
In contrast, LDVLMs employ bidirectional attention and iterative parallel unmasking, resulting in fundamentally different attention dynamics.
Consequently, existing rotary scaling methods do not directly address the attention imbalance and visual grounding challenges that arise in LDVLMs.
The empirical comparison with NTK and YaRN is reported in~\cref{tab:rope_baseline} of the main paper, where these baselines often degrade visual grounding while our monotonic low-frequency amplification yields consistent gains across both backbones.

\section{Comparison with Decoding-Level Baselines}
\label{sec:app_dpad}
We further compare our method against a decoding-level intervention that targets suffix-mask redundancy directly during inference.
We adopt a DPad-style suffix-dropping baseline, which is training-free and modifies suffix attention at the input level, making it directly applicable to existing LDVLM checkpoints.
This comparison is informative because DPad and our method address related but distinct mechanisms:
DPad reduces redundant suffix-mask influence at the input level, whereas our Monotonic RoPE Scaling strengthens long-range access to informative visual tokens at the representation level.

\cref{tab:app_dpad} shows that our method consistently outperforms DPad alone on both backbones across all three benchmarks.
On LaViDa, DPad and our method are largely complementary: combining them yields the best overall results.
On LLaDA-V, the combination is mixed: DPad's input-level masking may partially remove positional context that our Monotonic RoPE Scaling exploits, which limits additivity in this setting.

\begin{table}[t]
\centering
\small
\caption{\textbf{Comparison with a DPad-style decoding-level baseline.}
Our method is complementary to DPad on LaViDa, where the combination yields the best results across all three benchmarks.}
\begin{tabular}{lccc|ccc}
\toprule
& \multicolumn{3}{c|}{LLaDA-V} & \multicolumn{3}{c}{LaViDa} \\
\cmidrule(lr){2-4} \cmidrule(lr){5-7}
Method & RefCOCOg & Ferret & DetailCaps & RefCOCOg & Ferret & DetailCaps \\
\midrule
Baseline       & 64.8          & 60.4          & 59.8          & 36.9          & 25.9          & 8.3 \\
+ DPad         & 64.7          & 60.3          & 60.2          & 39.6          & 30.8          & 7.9 \\
+ Ours         & \textbf{65.0} & \textbf{62.9} & \textbf{63.6} & 44.0          & 35.7          & 56.1 \\
\rowcolor{gray!30}
+ DPad + Ours  & \textbf{65.0} & 61.7          & 60.5          & \textbf{44.4} & \textbf{36.1} & \textbf{56.6} \\
\bottomrule
\end{tabular}
\label{tab:app_dpad}
\end{table}

\myparagraph{On D2F.}
D2F~(diffusion-to-feed-forward) replaces the decoding paradigm with a block-wise autoregressive--diffusion hybrid, realized through asymmetric distillation from a pretrained DLLM, rather than as a training-free inference-time modification.
Moreover, it is introduced for text-only DLLMs.
Adapting D2F to existing LDVLM checkpoints would therefore require additional multimodal adaptation or training, which falls outside the scope of our training-free analysis.
We view it as a complementary direction and an interesting target for future work.

\section{Baselines}
\label{sec:baseline}
\myparagraph{LLaDA-V.}~\cite{you2025llada}
LLaDA-V adopts the LDVLM framework, building on the LLaDA~\cite{nie2025large} architecture.
The model comprises a SigLIP2-SO400M-Patch14-384 vision encoder~\cite{tschannen2025siglip} and an LLaDA-8B-Instruct language model.
During training, response tokens in a multi-turn dialogue are randomly masked and predicted via bidirectional attention.
LLaDA-V follows a three-stage training paradigm.
In the first stage, the model is trained on the LLaVA-Pretrain~\cite{liu2023visual} dataset to align visual and linguistic representations.
In the second stage, visual instruction tuning is performed using the MAmmoTH-VL~\cite{guo2025mammoth} dataset to enhance multimodal understanding.
In the final stage, multimodal reasoning capabilities are further strengthened using the VisualWebInstruct~\cite{jia2025visualwebinstruct} dataset, which contains approximately 900K question–answer pairs involving complex visual reasoning tasks.
Overall, LLaDA-V demonstrates performance comparable to autoregressive VLMs, highlighting its potential as an alternative modeling paradigm.

\myparagraph{LaViDa.}~\cite{li2025LaViDa}
LaViDa is an LDVLM proposed to address limitations of autoregressive MLLMs, particularly their slow inference speed and limited controllability during generation.
LaViDa employs a SigLIP-400M vision encoder \cite{zhai2023sigmoid} and an LLM backbone based on LLaDA-8B or Dream-7B \cite{ye2025dream}. 
In our experiments, we use LaViDa-L only, as it shares the same language backbone as LLaDA-V, enabling a fair comparison.
LaViDa introduces a complementary masking strategy during training. 
Instead of learning from a single masked version of a response, two complementary masked variants are constructed for each sample such that the masked regions do not overlap. 
This design ensures that all tokens contribute to training across iterations, alleviating efficiency and performance degradation caused by sparse masking.
LaViDa follows a two-stage training paradigm.
In the first stage, the model is pretrained by updating only the MLP projector to align visual embeddings with the LLM's word embedding space, using the LCS-558K dataset of 558K image–text pairs.
In the second stage, the entire model, including the vision encoder and the LLM, is jointly finetuned on approximately one million samples collected from multiple datasets such as COCO and ALLaVA-VFLAN \cite{chen2024allava}.
During inference, LaViDa further proposes Prefix-DLM, an inference optimization technique that caches and reuses the key–value states of prefix tokens, including visual tokens and prompt tokens. 
This caching mechanism reduces redundant computation and improves inference speed. 
In addition, LaViDa replaces the commonly used linear unmasking schedule with a shifting schedule, leading to further performance improvements.

\section{Additional Experiments on Unified Models}
\label{sec:app_unified}
We further apply our method to a unified model architecture.
Both models are evaluated using VLMEvalKit~\cite{duan2024vlmevalkit} under a consistent evaluation protocol.
For MMaDA~\cite{yang2025mmada}, we set $\lambda=0.1$, $\beta=0.4$, $k=3$, $\eta=8.0$, and $\tau_0=0.6$.
For Lumina-DiMOO~\cite{xin2025lumina},  we set $\lambda=0.1$, $\beta=0.4$, $k=3$, $\eta=12.0$, and $\tau_0=0.6$.
In both cases, our method consistently outperforms the corresponding baselines, as shown in~\cref{tab:app_unified}.
These results demonstrate that our approach generalizes beyond LDVLMs and can be effectively applied to unified vision–language architectures.
\begin{table}[t]
\centering
\small
\caption{\textbf{Quantitative Results on Unified Models}
Lumina refers to Lumina-DiMOO.
}
\begin{tabular}{lccc}
\toprule
 & MME-P & MMMU & MMB \\
\midrule

MMaDA (reported)
& 1410.7 & 30.2 & 68.5 \\
MMaDA (reproduced)
&1070.6  & 31.1 & 44.5   \\

\rowcolor{gray!20}
+Ours
&1117.2  & 32.7 & {44.5}  \\

\midrule
Lumina (reported)
& 1534.2 & 58.6 & 84.5  \\
Lumina (reproduced)
& 1543.6 & 59.3 & 84.9  \\
\rowcolor{gray!20}
+Ours
& {1553.1} & 62.0 & 85.0    \\

\bottomrule
\end{tabular}

\label{tab:app_unified}
\end{table}

\section{Additional Analysis and results on LaViDa and LLaDA-V}
\label{sec:app_analysis2}

\subsection{Analysis and Results on LaViDa}
\label{sec:lavida_rersult1}
\begin{figure}[t]
\includegraphics[width=\textwidth]
{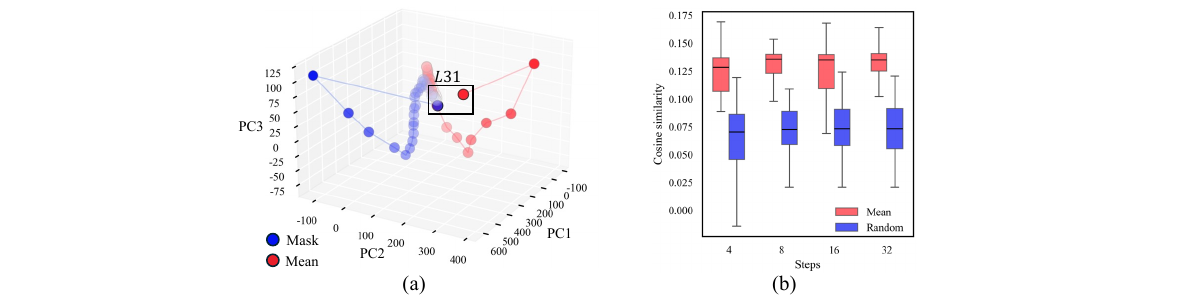}
    \caption{
    \textbf{Visualization of mask prior drift on LaViDa.}
    (a) 3D PCA trajectories of hidden states for the vocabulary mean embedding and the uncontextualized mask token, which converge to a similar region at the final layer ($L31$).
    (b) Cosine similarity between contextualized mask token embeddings and the vocabulary mean,
    showing consistently stronger alignment than random embeddings, especially with fewer generation steps.
    }
\label{fig:app_analysis1}
\end{figure}

\begin{figure}[t]
\includegraphics[width=\textwidth]
{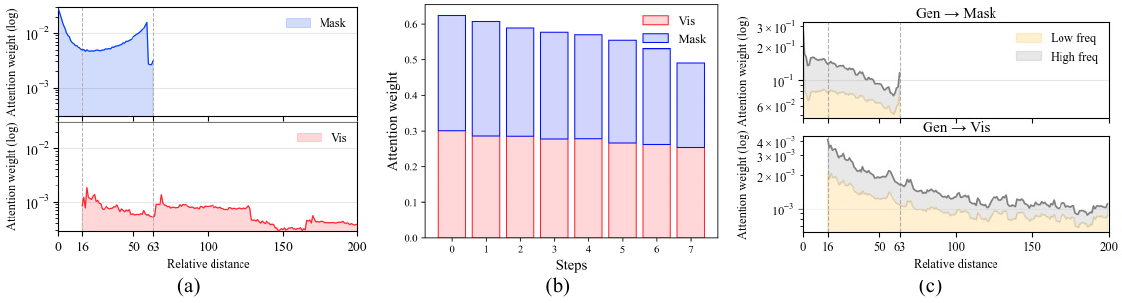}
\caption{
\textbf{Visualization of Positional Attention Collapse on LaViDa}
(a) Mean attention weight across relative distance (log scale), showing stronger attention to mask tokens than visual tokens at similar distances and a monotonic decay for visual tokens.
(b) Sum of attention to visual and mask tokens per generation token across generation steps, revealing a persistent allocation of comparable attention weights to mask tokens despite their lack of semantic content.
(c) Frequency decomposition of attention over relative distance, with high-frequency dominance at short ranges and low-frequency dominance at long ranges, yet weak overall attention.
}
\label{fig:app_freq}
\end{figure}

\begin{figure}[t]
\includegraphics[width=\textwidth]
{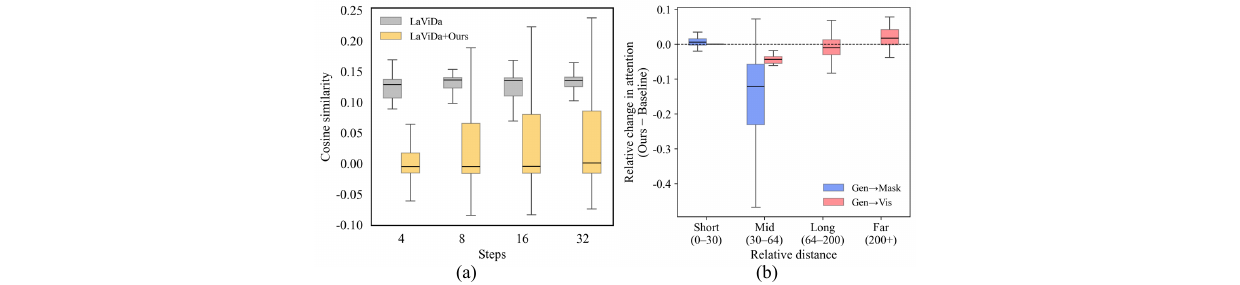}
\caption{
\textbf{Visualization of result analysis on LaViDa.}
(a) Box plot of cosine similarity between contextualized mask tokens and the vocabulary mean, showing consistent reduction across generation steps.
(b) Relative change in attention with respect to relative distance, where attention to distant visual tokens increases compared to the baseline, while attention to mask tokens is preserved or reduced.
}
\label{fig:app_result}
\end{figure}

In this section, we provide additional visual analyses of LaViDa to further support the observations presented in the main paper.
Specifically, we extend the analysis of mask token drift and positional attention collapse to LaViDa architecture.
As shown in Figs.~\ref{fig:app_analysis1} and~\ref{fig:app_freq}, the qualitative trends remain highly consistent with our previous findings, confirming that these phenomena are inherent characteristics of LDVLMs rather than artifacts of a specific model instance.
Moreover, as illustrated in Fig.~\ref{fig:app_result}, LaViDa exhibits the same
behavior observed in~\cref{sec:result_analysis}, where the mask prior is effectively
mitigated, and attention to visual tokens is preserved even at long relative distances.

\subsection{Additional Results on LLaDA-V}
\begin{figure}[t]
\includegraphics[width=\textwidth]
{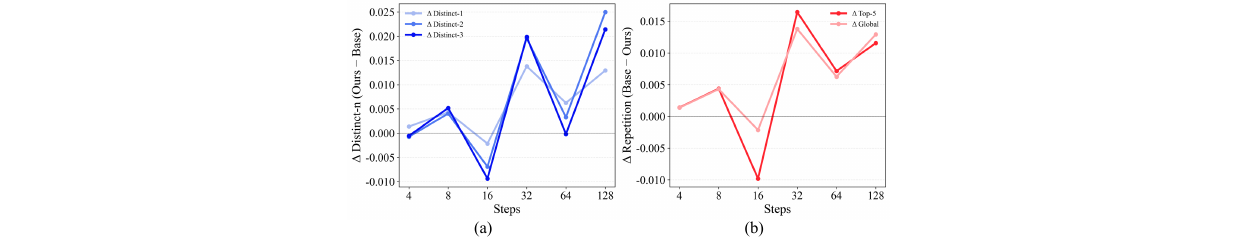}
\caption{
\textbf{Relative performance changes on DetailCaps across generation steps using LLaDA-V.}
Dashed lines: LLaDA-V, solid lines: Ours.
(a) $\Delta$Distinct-$n$ (Ours -- Base) shows consistent gains, with larger improvements at moderate to larger generation steps.
(b) $\Delta$Repetition ratio (Base -- Ours) remains positive across most steps, indicating reduced repetition, with the strongest reductions observed at intermediate steps.
}
\label{fig:app_quant}
\end{figure}

As shown in~\cref{fig:app_quant}, both distinct-$n$ and repetition ratios exhibit step-dependent improvements over the baseline,
with more pronounced gains observed at intermediate generation steps.

\subsection{Generation Step Analysis on DetailCaps}
\label{sec:lavida_rersult2}
\begin{figure}[t]
\includegraphics[width=\textwidth]
{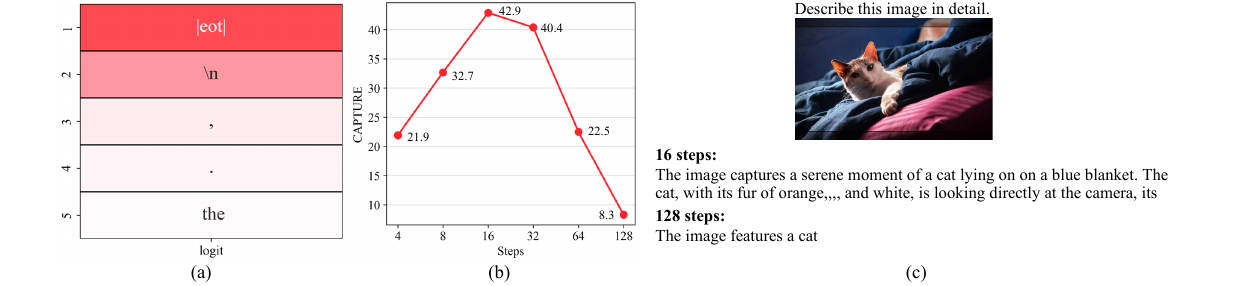}
\caption{
\textbf{Generation step analysis and DetailCaps performance on LaViDa.}
(a) Top-5 logits of the uncontextualized mask token $\mathcal{M}$, where the \texttt{|eot|} token consistently receives the highest logit.
(b) CAPTURE scores on the DetailCaps benchmark as a function of generation steps.
Contrary to a standard speed--quality trade-off, performance peaks at 16 steps and degrades with additional steps.
(c) Qualitative examples showing early termination induced by mask token prior at larger step counts (e.g., 32, 64, 128),
where frequent generation of \texttt{|eot|} results in prematurely truncated captions.
}
\label{fig:app_capture}
\vspace{-10pt}
\end{figure}

We further analyze the relationship between generation steps and DetailCaps performance on LaViDa.
As shown in~\cref{fig:app_capture}(a), the uncontextualized mask token $\mathcal{M}$ exhibits a strong bias toward the \texttt{|eot|} token,
which attains the highest logit among all vocabulary sets.
This indicates that initialized generation tokens as $\mathcal{M}$ inherently leads to early termination.
This bias is reflected in the quantitative results on the DetailCaps benchmark.
As illustrated in~\cref{fig:app_capture}(b), increasing the number of generation steps does not consistently improve caption quality.
Instead of following a conventional speed--quality trade-off,
the CAPTURE score peaks at 16 steps and subsequently decreases as the number of steps increases.
This behavior contrasts with the expectation that longer decoding should yield more detailed descriptions.
The underlying reason is closely related to mask prior drift.
As shown in~\cref{fig:app_capture}(c), when a large number of steps (e.g., 64, 128) is used,
the \texttt{|eot|} token is repeatedly generated across steps, causing captions to terminate prematurely.
In contrast, with a moderate number of steps,
more tokens are generated in parallel, reducing the dominance of the \texttt{|eot|} prior and mitigating early stopping.
This explains why LaViDa exhibits notably low DetailCaps performance in~\cref{tab:ablation},
and highlights the strong connection between mask prior drift and degraded long-form caption quality.

\section{Additional Experimental Results}
\label{sec:app_result}

\subsection{Reference-Based Captioning Metrics on RefCOCOg}
\label{sec:app_refmetrics}
We further report standard reference-based captioning metrics on RefCOCOg, which provides 7{,}573 images with human-annotated ground-truth references.
\cref{tab:app_refmetrics} shows that our method matches or improves upon the baselines across CIDEr, BLEU-4, METEOR, and RefCLIPScore.
For DetailCaps and LLaVA-Bench, standard reference-based metrics are less directly applicable: DetailCaps does not provide canonical reference captions, and LLaVA-Bench uses GPT-4 judge scores as its evaluation protocol.
We therefore report task-level scores in the main tables for these benchmarks.

\begin{table}[t]
\centering
\small
\caption{\textbf{Reference-based captioning metrics on RefCOCOg.}
Higher is better for all metrics.}
\begin{tabular}{lcccc}
\toprule
Method   & CIDEr         & BLEU-4        & METEOR        & RefCLIPScore  \\
\midrule
LLaDA-V  & 64.8          & 8.2           & 14.6          & 72.8 \\
\rowcolor{gray!30}
+ Ours   & \textbf{65.0} & \textbf{8.4}  & 14.6          & 72.8 \\
LaViDa   & 36.9          & 1.5           & 8.4           & 68.1 \\
\rowcolor{gray!30}
+ Ours   & \textbf{44.0} & \textbf{4.9}  & \textbf{11.3} & \textbf{69.4} \\
\bottomrule
\end{tabular}
\label{tab:app_refmetrics}
\end{table}

\subsection{Layer-wise Application of Mask Prior Suppression}
\label{sec:app_mps_layer}
Our default configuration applies Mask Prior Suppression (MPS) only at the final transformer layer.
We ablate this design by progressively expanding MPS to deeper subsets of layers.
\cref{tab:app_mps_layer} shows that applying MPS at the final layer alone provides the most stable trade-off across benchmarks and is consistently strongest on long-form generation (LLaVA-Bench, DetailCaps).
Suppressing the prior at earlier layers can interfere with intermediate representations that still encode useful contextual information, particularly hurting long-form generation on LaViDa.
The reported \emph{prior} column denotes the average ratio of suppressed prior energy across the configured layers; with a fixed total budget, applying MPS to more layers necessarily distributes per-layer suppression more thinly.

\begin{table}[t]
\centering
\small
\caption{\textbf{Layer-wise application of Mask Prior Suppression.}
\emph{Final-only} denotes our default setting.}
\setlength{\tabcolsep}{5pt}
\begin{tabular}{llcccc}
\toprule
Model & Configuration & Prior & RefCOCOg & LLaVA-Bench & DetailCaps \\
\midrule
\multirow{5}{*}{LLaDA-V}
& Baseline                  & --     & 64.8 & 61.3 & 59.8 \\
& All layers ($\times 32$)  & 0.003  & 61.3 & 63.5 & 60.2 \\
& Last 8 layers             & 0.013  & \textbf{65.1} & 61.1 & 59.8 \\
& Last 3 layers             & 0.033  & 64.8 & 61.1 & 60.3 \\
\rowcolor{gray!30}
& Final-only (Ours)         & 0.100  & 65.0 & \textbf{64.1} & \textbf{63.6} \\
\midrule
\multirow{5}{*}{LaViDa}
& Baseline                  & --     & 36.9 & 39.5 & 8.3 \\
& All layers ($\times 32$)  & 0.009  & 40.0 & 37.4 & 6.7 \\
& Last 8 layers             & 0.038  & 39.6 & 36.7 & 8.1 \\
& Last 3 layers             & 0.100  & 39.7 & 35.8 & 8.1 \\
\rowcolor{gray!30}
& Final-only (Ours)         & 0.300  & \textbf{44.0} & \textbf{46.5} & \textbf{56.1} \\
\bottomrule
\end{tabular}
\label{tab:app_mps_layer}
\end{table}

\subsection{Segment-wise Application of Monotonic RoPE Scaling}
\label{sec:app_mrs_segment}
A natural concern is that frequency rescaling might distort 2D visual spatial structure when applied uniformly across visual, textual, and generation tokens.
Although Monotonic RoPE Scaling rescales RoPE frequency components without changing token indices, patch layout, or sequence order, we provide a direct empirical check by restricting the rescaling to specific token segments.
\cref{tab:app_mrs_segment} reports the results when MRS is applied only to visual tokens, only to textual prompt tokens, only to generation tokens, or to all tokens (default).

A few observations follow.
First, applying MRS to visual tokens alone already improves visual grounding on both backbones, indicating that the rescaling does not degrade 2D visual spatial structure.
Second, no single segment consistently dominates: on LLaDA-V, applying MRS to all tokens is best on DetailCaps; on LaViDa, applying MRS to either visual or textual tokens alone is best on DetailCaps but very close to the all-tokens setting on the remaining benchmarks.
Overall, the uniform all-tokens setting provides the strongest or near-strongest performance across benchmarks without task-specific tuning, which we adopt as the default.

\begin{table}[t]
\centering
\small
\caption{\textbf{Segment-wise application of Monotonic RoPE Scaling.}
``vis only'', ``txt only'', and ``gen only'' restrict MRS to visual, textual, or generation tokens; ``all'' is the default.}
\setlength{\tabcolsep}{5pt}
\begin{tabular}{llccc}
\toprule
Backbone & Variant     & RefCOCOg      & Ferret        & DetailCaps \\
\midrule
\multirow{5}{*}{LLaDA-V}
& Baseline      & 64.8          & 60.4          & 59.8 \\
& vis only      & \textbf{65.2} & 62.5          & 59.6 \\
& txt only      & 65.0          & \textbf{63.2} & 60.2 \\
& gen only      & 64.9          & 61.8          & 60.1 \\
\rowcolor{gray!30}
& all (Ours)    & 65.0          & 62.9          & \textbf{63.6} \\
\midrule
\multirow{5}{*}{LaViDa}
& Baseline      & 36.9          & 25.9          & 8.3 \\
& vis only      & 43.8          & 35.9          & \textbf{56.4} \\
& txt only      & 43.8          & \textbf{36.1} & \textbf{56.4} \\
& gen only      & 43.8          & 35.8          & \textbf{56.4} \\
\rowcolor{gray!30}
& all (Ours)    & \textbf{44.0} & 35.7          & 56.1 \\
\bottomrule
\end{tabular}
\label{tab:app_mrs_segment}
\end{table}

\subsection{Hyperparameter Sensitivity}
\label{sec:app_hyper}
We analyze the hyperparameter sensitivity of our method to assess its robustness.
We evaluate the impact of five hyperparameters:
$\lambda$ for prior suppression strength,
$\beta$ for RoPE scaling magnitude,
$k$ for prior subspace dimensionality,
$\eta$ for the slope of monotonic RoPE scaling,
and $\tau_0$ for the center of the frequency-wise scaling setting.
\cref{tab:app_hyper_lladav} and~\cref{tab:app_hyper_lavida} summarize the results across multiple benchmarks.
Overall, we observe that performance varies smoothly with respect to $\lambda$ and $\beta$,
indicating low sensitivity over a broad range of values.
In contrast, the choice of $k$ exhibits a clear trade-off between expressiveness and stability:
very small values limit the capacity of prior suppression, while excessively large values lead to unstable behavior,
especially in long-form generation.
Moderate values such as $k=3$ consistently provide a favorable balance across benchmarks.
We further find that the method is relatively robust to the RoPE scaling parameters $\eta$ and $\tau_0$.
Varying $\eta$ mainly affects the sharpness of frequency-wise reweighting, with moderate slopes yielding stable performance,
while extreme values provide limited additional benefits.
Similarly, adjusting $\tau_0$ results in only minor performance fluctuations,
suggesting that the method does not rely on precise tuning of the frequency center.
These observations indicate that the proposed monotonic RoPE scaling is stable across a wide range of configurations.

\begin{table}[t]
\centering
\small
\caption{\textbf{Hyperparameter sensitivity analysis on LLaDA-V.}
The highlighted row denotes the default configuration used in all main experiments.}
\setlength{\tabcolsep}{5pt}
\begin{tabular}{ccccc|cc|cc}
\toprule
\multicolumn{5}{c|}{Hyperparameter}
& \multicolumn{2}{c|}{Visual Grounding}
& \multicolumn{2}{c}{Long-form Generation} \\
\cmidrule(lr){1-5}
\cmidrule(lr){6-7}
\cmidrule(lr){8-9}

$\lambda$ & $\beta$ & $k$ & $\eta$ & $\tau_0$
& RefCOCOg & Ferret
& LLaVA-Bench & DetailCaps \\
\midrule

0.2 & 0.1 & 3 & 8 & 0.6
& 64.8 & 62.4 & 61.9 & 60.0 \\

0.3 & 0.1 & 3 & 8 & 0.6
& 64.2 & 61.3 & 64.7 & 60.4 \\
\midrule
0.1 & 0.2 & 3 & 8 & 0.6
& 65.3 & 61.8 & 59.8 & 66.4 \\

0.1 & 0.3 & 3 & 8 & 0.6
& 65.2 & 61.7 & 63.9 & 59.7 \\

\midrule
0.1 & 0.1 & 1 & 8 & 0.6
& 66.9 & 60.7 & 60.1 & 60.1 \\

0.1 & 0.1 & 8 & 8 & 0.6
& 64.4 & 60.3 & 61.8 & 60.0 \\

0.1 & 0.1 & 16 & 8 & 0.6
& 64.5 & 61.7 & 62.5 & 60.1 \\

\midrule
0.1 & 0.1 & 3 & 4 & 0.6
& 64.8 & 61.1 & 60.5 & 59.5 \\

0.1 & 0.1 & 3 & 10 & 0.6
& 64.9 & 61.1 & 63.7 & 60.0 \\

\midrule
0.1 & 0.1 & 3 & 8 & 0.4
& 65.3 & 61.2 & 61.9 & 60.0 \\

0.1 & 0.1 & 3 & 8 & 0.8
& 64.9 & 62.3 & 64.0 & 59.8 \\

\midrule
\rowcolor{gray!30}
0.1 & 0.1 & 3 & 8 & 0.6
& 65.0 & {62.9} & {64.1} & {63.6} \\

\bottomrule
\end{tabular}
\label{tab:app_hyper_lladav}
\vspace{-10pt}
\end{table}

\begin{table}[t]
\centering
\small
\caption{\textbf{Hyperparameter sensitivity analysis on LaViDa.}
The highlighted row denotes the default configuration used in all main experiments.}
\setlength{\tabcolsep}{5pt}
\renewcommand{\arraystretch}{1.1}
\begin{tabular}{ccccc|cc|cc}
\toprule
\multicolumn{5}{c|}{Hyperparameter}
& \multicolumn{2}{c|}{Visual Grounding}
& \multicolumn{2}{c}{Long-form Generation} \\
\cmidrule(lr){1-5}
\cmidrule(lr){6-7}
\cmidrule(lr){8-9}

$\lambda$ & $\beta$ & $k$ & $\eta$ & $\tau_0$
& RefCOCOg & Ferret
& LLaVA-Bench & DetailCaps \\
\midrule

0.2 & 0.1 & 3 & 12 & 0.6
& 44.0 & 34.6
& 46.3 & 48.4 \\
0.4 & 0.1 & 3 & 12 & 0.6
& 44.0 & 43.2
& 43.2 & 57.1 \\

\midrule
0.3 & 0.2 & 3 & 12 & 0.6
& 44.0 & 35.0
& 40.8 & 56.1 \\
0.3 & 0.3 & 3 & 12 & 0.6
& 44.0 & 34.7
& 45.6 & 56.1 \\

\midrule
0.3 & 0.1 & 1 & 12 & 0.6
& 39.9 & 29.3
& 41.1 & 8.3 \\
0.3 & 0.1 & 8 & 12 & 0.6
& 44.7 & 38.4
& 47.8 & 56.8 \\
0.3 & 0.1 & 16 & 12 & 0.6
& 44.8 & 36.9
& 38.1 & 56.9 \\

\midrule
0.3 & 0.1 & 3 & 10 & 0.6
& 44.0 & 35.3
& 44.9 & 56.1 \\
0.3 & 0.1 & 3 & 14 & 0.6
& 44.0 & 34.7
& 47.3 & 56.1 \\

\midrule
0.3 & 0.1 & 3 & 12 & 0.4
& 44.0 & 35.3
& 43.4 & 56.1 \\
0.3 & 0.1 & 3 & 12 & 0.8
& 44.0 & 35.0
& 44.7 & 56.1 \\
\midrule
\rowcolor{gray!30}
0.3 & 0.1 & 3 & 12 & 0.6
& {44.0} & {35.7}
& {46.5} & {56.1} \\

\bottomrule
\end{tabular}
\label{tab:app_hyper_lavida}
\vspace{-10pt}
\end{table}

\subsection{Inference latency}
\label{sec:latency}
We evaluate the inference latency of our method under varying numbers of generation steps.
As shown in Table~\ref{tab:app_latency}, our approach introduces a small and consistent latency overhead compared to the base models.
The additional latency scales proportionally with the number of generation steps, indicating that the proposed modifications preserve the efficiency of the underlying decoding process.
\begin{table}[t]
\centering
\small
\caption{
\textbf{Latency under different generation steps.}
We measure inference latency by varying the number of generation steps while fixing the generation length to 64.
}
\setlength{\tabcolsep}{6pt}
\begin{tabular}{l*{4}{c}}
\toprule
Model & \multicolumn{4}{c}{Generation Steps} \\
\cmidrule(lr){2-5}
      & 4 & 8 & 16 & 32 \\
\midrule
LLaDA-V & 2.31s & 4.56s & 9.07s & 18.10s \\
\rowcolor{gray!30}
+ Ours  & 2.33s & 4.60s & 9.11s & 18.16s \\
\midrule
LaViDa  & 0.49s & 0.92s & 1.79s & 3.52s \\
\rowcolor{gray!30}
+ Ours  & 0.57s & 1.09s & 2.12s & 4.19s \\
\bottomrule
\end{tabular}
\label{tab:app_latency}
\end{table}

\subsection{RoPE Scheduling}
\vspace{-5pt}
We study the effect of different RoPE scheduling functions on visual grounding and long-form generation performance.
\cref{tab:app_schedule} compares sigmoid, cosine, exponential, linear, and power schedules under the same setting.
Across both LLaDA-V and LaViDa, we observe that the choice of scheduling function has a more pronounced impact on visual grounding than on long-form generation.
In particular, the sigmoid schedule consistently yields strong performance on visual grounding benchmarks such as RefCOCOg and Ferret.
Compared to alternative schedules, sigmoid provides stable improvements without introducing performance degradation across tasks.
We attribute this behavior to the smooth and monotonic nature of the sigmoid schedule, which gradually modulates low-frequency positional components while preserving high-frequency information.
This property helps maintain long-range visual–textual alignment, which is critical for visual grounding tasks.
In contrast, schedules with more abrupt or aggressive scaling, such as exponential or power functions, can distort positional biases and lead to less stable grounding performance.
Overall, these results suggest that smooth, monotonic scheduling functions are better suited for controlling RoPE-induced locality bias in multimodal grounding scenarios.
\vspace{-5pt}
\begin{table}[t]
\centering
\small
\caption{\textbf{Ablation study on different scheduling functions.}
We compare sigmoid, cosine, exponential, linear, and power schedules
on visual grounding and long-form generation benchmarks.\textbf{Bold}: best, \underline{underline}: second best.
}
\begin{tabular}{llcccc}
\toprule
\multirow{2}{*}{Model}
& \multirow{2}{*}{Type}
& \multicolumn{2}{c}{Visual Grounding}
& \multicolumn{2}{c}{Long-form Generation} \\
\cmidrule(lr){3-4}
\cmidrule(lr){5-6}

& 
& RefCOCOg
& Ferret
& LLaVA-Bench
& DetailCaps \\
\midrule

\multirow{5}{*}{LLaDA-V}

& Cosine
& \textbf{65.1} & \textbf{63.4} & \underline{62.7} & 60.6 \\
& Exp
& \underline{65.0} & 61.4 & 61.4 & 60.2 \\
& Linear
& \textbf{65.1} & 61.3 & 62.2 & \underline{60.7} \\
& Power
& 64.9 & 63.5 & 61.5 & 60.3 \\

& \cellcolor{gray!30}\textbf{Sigmoid}
& \cellcolor{gray!30}\underline{65.0}
& \cellcolor{gray!30}\underline{62.9}
& \cellcolor{gray!30}\textbf{64.1}
& \cellcolor{gray!30}\textbf{63.6} \\

\midrule

\multirow{5}{*}{LaViDa}

& Cosine
& 43.8 & 34.3 & 45.9 & \textbf{56.6} \\

& Exp
& \underline{44.0} & 34.1 & \textbf{46.8} & \underline{56.4} \\

& Linear
& 43.8 & 32.6 & 48.0 & 48.9 \\

& Power
& \textbf{44.2} & \underline{34.3} & 43.3 & 55.5 \\

& \cellcolor{gray!30}\textbf{Sigmoid}
& \cellcolor{gray!30}\underline{44.0}
& \cellcolor{gray!30}\textbf{35.7}
& \cellcolor{gray!30}\underline{46.5}
& \cellcolor{gray!30}{56.1} \\
\bottomrule
\end{tabular}
\vspace{-10pt}
\label{tab:app_schedule}
\end{table}

\section{Qualitative Results}
\label{sec:app_qual}
\begin{figure}[t]
\includegraphics[width=\textwidth]
{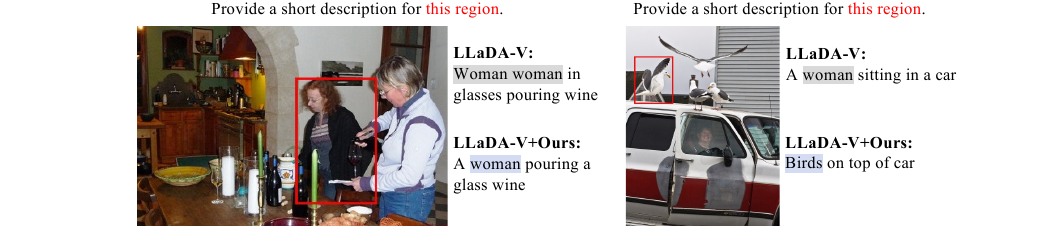}
\caption{\textbf{Qualitative results on RefCOCOg using LLaDA-V.}
The red bounding boxes indicate the target regions in the image.
}
\label{fig:app_lladav_rg}
\end{figure}

\begin{figure}[t]
\includegraphics[width=\textwidth]
{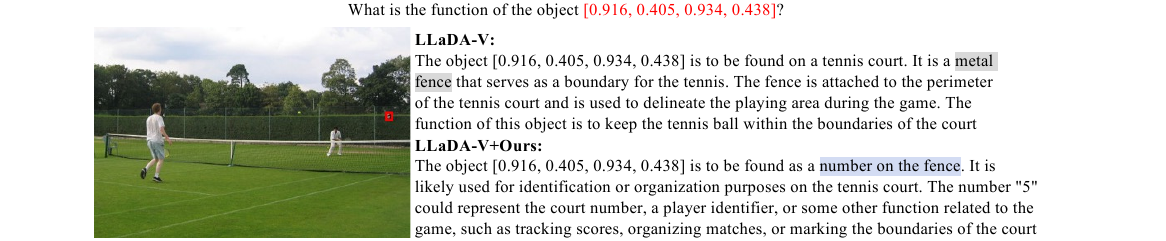}
\caption{\textbf{Qualitative results on Ferret using LLaDA-V.}
The red bounding boxes indicate the target regions in the image.
}
\label{fig:app_lladav_ft}
\end{figure}

\begin{figure}[t]
\includegraphics[width=\textwidth]
{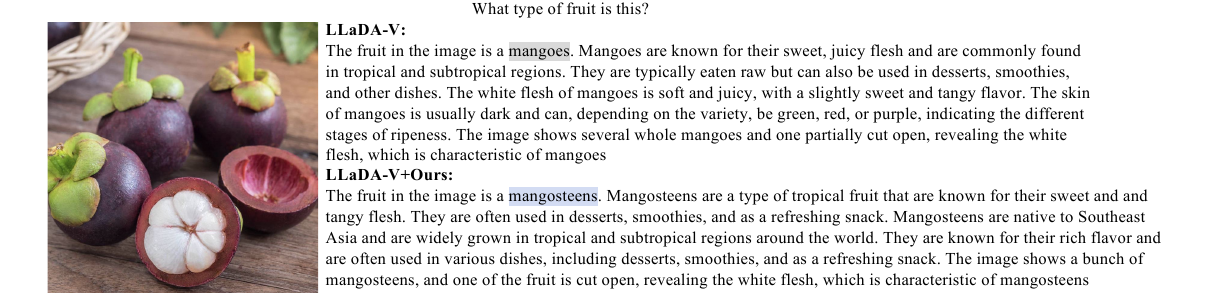}
\caption{\textbf{Qualitative results on LLaVA-Bench using LLaDA-V.}
}
\label{fig:app_lladav_lb}
\end{figure}

\begin{figure}[t]
\includegraphics[width=\textwidth]
{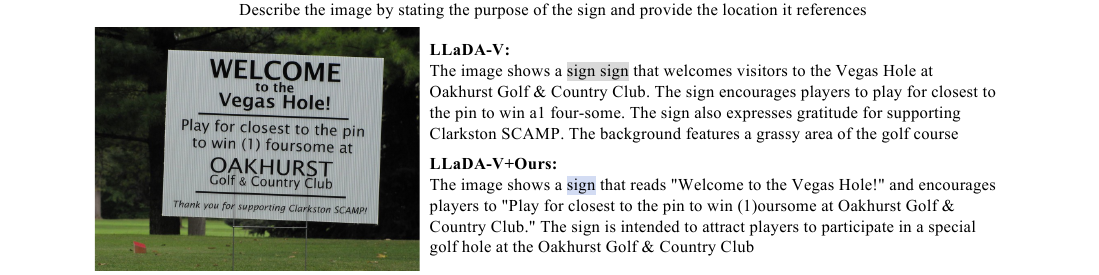}
\caption{\textbf{Qualitative results on MIA using LLaDA-V.}
}
\label{fig:app_lladav_mb}
\end{figure}

\begin{figure}[t]
\includegraphics[width=\textwidth]
{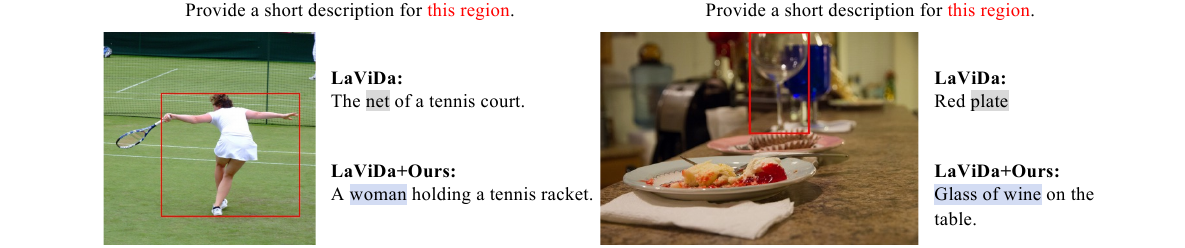}
\caption{\textbf{Qualitative results on RefCOCOg using LaViDa.}
The red bounding boxes indicate the target regions in the image.
}
\label{fig:app_rg}
\end{figure}

\begin{figure}[t]
\includegraphics[width=\textwidth]
{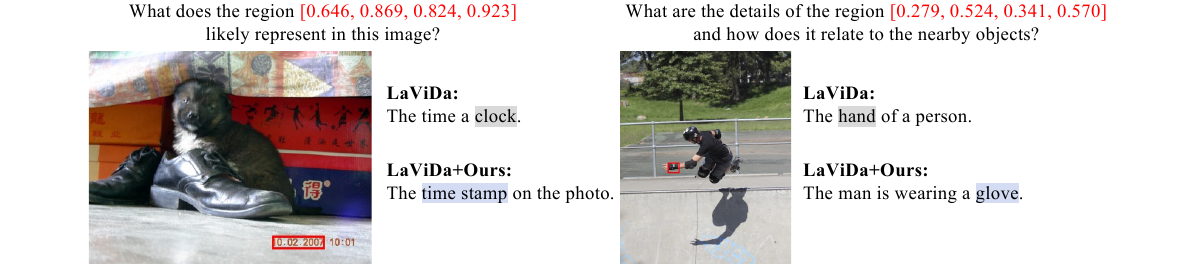}
\caption{\textbf{Qualitative results on Ferret using LaViDa.}
The red bounding boxes indicate the target regions in the image.
}
\label{fig:app_ft}
\end{figure}

\begin{figure}[t]
\includegraphics[width=\textwidth]
{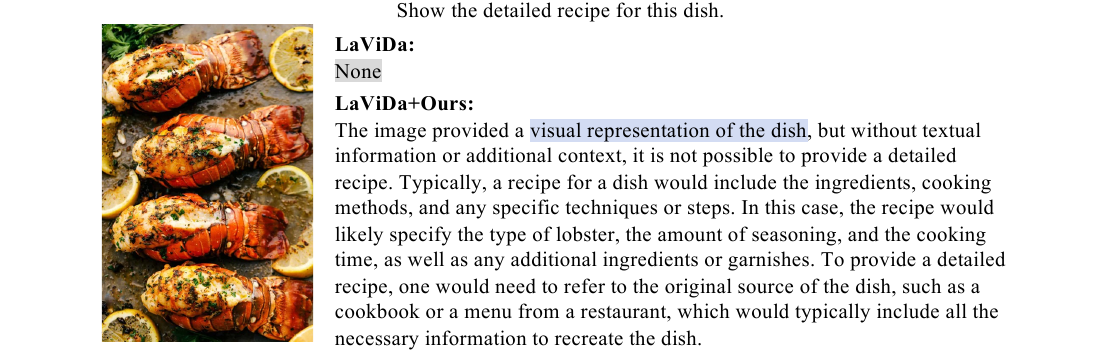}
\caption{\textbf{Qualitative results on LLaVA-Bench using LaViDa.}
}
\label{fig:app_lb}
\end{figure}

\begin{figure}[t]
\includegraphics[width=\textwidth]
{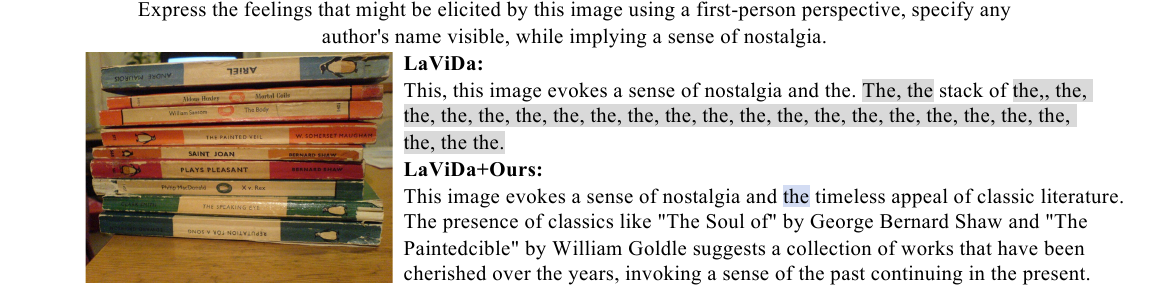}
\caption{\textbf{Qualitative results on MIA using LaViDa.}
}
\label{fig:app_mb}
\end{figure}

Additional qualitative results on visual grounding and long-form generation for LLaDA-V are provided in
\cref{fig:app_lladav_rg,fig:app_lladav_ft,fig:app_lladav_lb,fig:app_lladav_mb}.
Additional qualitative results for LaViDa are provided in
\cref{fig:app_rg,fig:app_ft,fig:app_lb,fig:app_mb}.

\section{Limitations and Future Work}
\label{sec:app_limitation}
\myparagraph{Limitations.}
While our method consistently improves visual grounding and long-form generation without additional training, it is primarily designed for inference-time intervention and does not modify the underlying model parameters.
As a result, the effectiveness of our approach depends on the quality of the pretrained representations and may be limited when applied to models with weaker multimodal alignment or different positional encoding schemes.
In addition, our analysis focuses on attention dynamics and hidden-state behavior in masked diffusion-based vision-language models, and the generalization of these findings to other generation paradigms remains to be fully explored.
We further note that our experiments are conducted under the standard full-suffix iterative unmasking setup adopted by current LDVLMs.
The DPad comparison in~\cref{sec:app_dpad} indicates that our methods remain effective when suffix redundancy is partially reduced;
however, validating our approach in more substantially modified decoding paradigms—such as semi-autoregressive or block-wise diffusion variants like D2F, which require additional multimodal adaptation—remains an important direction for future work.

\myparagraph{Future work.}
While this study establishes the efficacy of the proposed mechanisms, several promising directions remain for future investigation.
First, we aim to integrate these mechanisms into training-time optimization in order to examine their effects on model convergence and the formation of intrinsic representations.
Second, we plan to extend our analysis to alternative positional encoding schemes, including various variants of RoPE and their applications to LDVLMs, to gain a deeper understanding of the generality of our findings.
Finally, evaluating the robustness of our approach in complex multimodal settings, such as video-language understanding and long-context multi-turn generation, constitutes an important step toward more general and scalable multimodal systems.

\end{document}